\documentclass[lettersize,journal]{IEEEtran}
\usepackage{amsmath,amsfonts}
\usepackage{algorithmic}
\usepackage{algorithm}
\usepackage{array}
\usepackage[caption=false,font=normalsize,labelfont=sf,textfont=sf]{subfig}
\usepackage{textcomp}
\usepackage{stfloats}
\usepackage{url}
\usepackage{verbatim}
\usepackage{graphicx}
\usepackage{cite}
\usepackage{booktabs}
\usepackage{multirow}
\begin{document}

\title{The Neuro-Physical Inverter: A Modular Framework for Magnetotelluric Inversion Coupling Ensemble Conditioning with Residual Learning}

\author{Jae Deok Kim, Sai Ravela, Rob. L. Evans 

\thanks{Jae Deok Kim was with the MIT-WHOI Joint Program in Oceanography/Applied Ocean Science and Engineering, Cambridge, MA 02139 USA and is now with the Department of Earth, Atmospheric and Planetary Sciences, Massachusetts Institute of Technology, Cambridge, MA 02139 USA (e-mail: kimjd@mit.edu).}
\thanks{Sai Ravela is with the Department of Earth, Atmospheric and Planetary Sciences, Massachusetts Institute of Technology, Cambridge, MA 02139 USA.} 
\thanks{Rob. L. Evans is with the Department of Geology and Geophysics, Woods Hole Oceanographic Institution, Woods Hole, MA 02543 USA.}
}

\markboth{IEEE TRANSACTIONS ON GEOSCIENCE AND REMOTE SENSING}%
{}

\maketitle

\begin{abstract}
We present the Neuro-Physical Inverter (NPI), a modular, uncertainty-aware framework for geophysical inversion that couples ensemble-based conditioning with constrained residual learning, demonstrated in the 1D magnetotelluric (MT) setting as a controlled testbed. The framework operates in two stages. An Ensemble-Conditional Gaussian Process (EnsCGP) conditions a prior ensemble of resistivity models on the observed response, producing a physically admissible reference ensemble. A residual-learning neural network then predicts targeted corrections to this reference, trained on synthetic data and fine-tuned per station for field application through a physics-coupled objective. Because an ensemble is conditioned, refined, and propagated through both stages, every estimate carries an associated ensemble spread. Synthetic experiments show that NPI systematically reduces ensemble-mean error without destabilizing the ensemble. Applied to broadband MT data from the Gabbs Valley geothermal region (Nevada, USA), NPI reduces the across-station mean misfit over the mid-period band while retaining comparable ensemble spread. The propagated ensemble yields a factor of uncertainty that serves as an operational measure of constraint within the assumed model class. Both stages are dimension-agnostic in formulation, and the design principles established here are intended to scale to higher-dimensional parameterizations.
\end{abstract}


\section{Introduction}
\label{sec:intro}
\IEEEPARstart{G}{eophysical} methods estimate subsurface physical properties from indirect measurements made at or near the Earth's surface \cite{tarantola_ch1}. Such inverse problems are typically ill-posed. Different subsurface structures can produce similar observations, and noise or incomplete spatial and frequency coverage further complicate interpretation \cite{aster_ch3}. Magnetotellurics (MT), a passive electromagnetic (EM) method, exemplifies these challenges. MT records naturally varying electric and magnetic fields to estimate subsurface electrical resistivity (or its inverse, conductivity) \cite{Chave_Jones_2012_ch1}, a parameter linked to temperature, fluids, and composition \cite{Evans_2012}. Although the MT forward response is uniquely determined by the resistivity model, the inverse problem is strongly non-unique and noise-sensitive \cite{aster_ch3}, so reliable interpretation requires inversion methods that are both robust and uncertainty-aware.

For 1D MT, the inverse problem is relatively well studied. Deterministic methods such as Occam inversion \cite{constable_occam_1987} and nonlinear conjugate-gradient schemes \cite{rodi_mackie_nlcg_2001} provide efficient regularized solutions, and probabilistic methods such as Markov chain Monte Carlo (MCMC) \cite{tarantola_ch2, blatter_2019, blatter_part1_2022} give rigorous posterior characterization at modest cost. We do not aim to replace these. Our target is the higher-dimensional regime that 1D stands in for, where these classes face sharper cost, regularization, and reliability trade-offs. This motivates building an inversion framework from components that remain tractable as dimensionality grows. Two natural candidates, ensemble-based conditioning and data-driven learning, are attractive for this reason but each carries a characteristic failure mode.

Ensemble-based methods represent variability through a finite ensemble of realizations updated by derivative-free, covariance-based adjustments \cite{evensen_EnsembleKalmanFilter_2003, Evensen_2009}, which scale favorably and stay within a physically admissible model class. They face two limitations: a single linear update confined to the prior ensemble span cannot represent structured misfit lying outside that span, and real MT data can deviate from the statistical structure of the prior. This deviation, through cultural noise (e.g., electromagnetic interference from power lines, electrified railways, pipeline cathodic-protection currents), incomplete frequency coverage, or unmodeled site effects \cite{egbert_1997, Ferguson_2012, Simpson_Bahr_2005_ch3}, can destabilize or collapse the update unless it is stabilized \cite{Chen_2013, emerick_2013}. We therefore use ensemble updating as a single-pass conditioning operator rather than as an iterative solver, distinguishing it from iterative ensemble inversion methods (e.g., ensemble Kalman inversion, ES-MDA) that reuse observations across iterations as a derivative-free optimization strategy; this distinction is developed in Section~\ref{sec:enscgp_vs_eki}.

Machine learning has been widely applied to geophysical inversion, including data-driven estimation of resistivity from EM measurements \cite{puzyrev_2019, puzyrev_2021, liu_2021, wu_2023, xu_2024} and strategies for embedding physical constraints, such as physics-informed losses \cite{raissi_2019_pinn, daw_2021_pgnn} and residual convolutional architectures for MT \cite{liu_DeepLearningAudio_2021}. A related line of work targets uncertainty directly, training networks to predict posterior statistics from EM data \cite{hansen_2022}, using invertible architectures to draw posterior samples \cite{bittar_InvertibleNeuralNetwork_2026}, or applying variational and ensemble-based generators to frequency-domain EM \cite{narciso_ComparisonKalmanEnsemble_2022, liu_FrequencydomainElectromagneticInduction_2023}. These networks predict rapidly once trained \cite{lecun_DeepLearning_2015}, but tend to overfit the synthetic training distribution and degrade under noise unlike that seen in training \cite{rudin_2019_stop_black_box_ML, li_2022_interpretable_DL}.

These two failure modes are complementary: ensemble conditioning is stable and physics-consistent but cannot reach misfit outside the prior span, while a learned corrector can capture that residual nonlinearity. Coupling them lets each cover the other's gap, within a modular framework whose components can be substituted independently as dimensionality increases. The inversion is uncertainty-aware by construction, with an ensemble propagated through both stages. The conditioning step also leaves the network only a low-variance, largely location-independent residual to learn. That is a claim about the tractability of the learning problem rather than about point-estimate accuracy, which a sufficiently expressive network trained on the full mapping can match (Section~\ref{sec:discussion_resnet}).

We develop and verify this framework in 1D MT. Using 1D as a controlled setting is well established in data-driven geophysical inversion, where the inexpensive forward operator and the availability of synthetic ground truth make it a standard testbed for developing and evaluating learned inversion schemes before extension to higher dimensions \cite{liu_DeepLearningAudio_2021, rahmanijevinani_DeepLearningbased1D_2024, goyes_2025, saibi_ComparisonDeepLearning_2026}. We adopt it in the same spirit, as a deliberate methodological choice rather than a constraint of the formulation.

We develop the Neuro-Physical Inverter (NPI), a modular inversion framework that couples an Ensemble-Conditional Gaussian Process (EnsCGP) with constrained residual learning. In the first stage, EnsCGP updates a prior ensemble of resistivity models using observed MT data through a stabilized, single-pass Kalman-type adjustment \cite{evensen_EnsembleKalmanFilter_2003, saha_2024, trautner_2020_informative_ensemble_Kalman_Learning, ravela_2007, ravela_EnsembleconditionalGaussianProcesses_2026}. The output is a physically admissible ensemble whose forward responses are broadly consistent with the observations. Because this update is linear and confined to the prior ensemble span, however, structured misfit can persist where the prior is insufficient or the forward operator is strongly nonlinear in the relevant model directions. The second stage addresses this with a residual neural network that predicts depth-dependent corrections to the EnsCGP ensemble mean from the model-consistent MT responses produced by EnsCGP. The network never maps observations directly to resistivity; it acts only as a constrained correction to an existing physically admissible reference. The framework is developed in 1D, but is not specific to it. The EnsCGP update depends only on model and response ensembles and their empirical covariances, so the same formal structure applies in 2D and 3D.

The two-stage decomposition simplifies learning: the network learns only the discrepancy between the EnsCGP baseline and the target (Section~\ref{sec:resnet}), following a broader pattern in data-driven scientific modeling where a physics-based baseline captures dominant structure and a learned corrector handles the residual \cite{dong2025preadaptfwipretrainedbasedadaptiveresidual, dong_Enhanced3DGravity_2026, mardani_ResidualCorrectiveDiffusion_2025, saha_2024, song_SemisupervisedLearningSeismic_2022}. What distinguishes NPI is that the baseline is itself physics-based, restricting the corrector to a physically admissible model class. The ensemble produced at each stage is propagated forward and reported as a workflow-derived measure of model variability, not a full posterior.

The main contributions of this work are:
\begin{itemize}
    \item \textbf{A dimension-agnostic, modular inversion framework (NPI)} coupling single-pass ensemble conditioning with constrained residual learning, designed for extension to large-scale, high-dimensional problems.

    \item \textbf{A formulation of EnsCGP as a conditioning operator}, distinct from iterative optimization or posterior sampling, that produces physically admissible first-guess estimates whose forward responses are consistent with observations.

    \item \textbf{A physics-constrained residual-learning strategy} that corrects structured misfit relative to the ensemble-conditioned baseline, operating on model-consistent responses and enforcing minimum-change refinement.

    \item \textbf{A controlled 1D demonstration} on synthetic data and on real broadband MT observations from the Gabbs Valley geothermal region (Nevada, USA), examining each component as a step toward higher-dimensional inversion.
\end{itemize}

\section{Methodology}
\label{sec:methods}

\subsection{Overview of the Neuro-Physical Inverter (NPI)}
\label{sec:overview_npi}
The NPI couples ensemble-based conditioning (EnsCGP, Section~\ref{sec:EnsCGP}) with constrained residual correction (ResNet-1D, Section~\ref{sec:resnet}). EnsCGP produces a physically admissible reference ensemble from the observed MT response, whose forward-modeled responses we term \emph{model-consistent} to emphasize that they derive from a specific model rather than direct observation. The residual corrector then predicts depth-dependent corrections to that reference, addressing structured misfit the ensemble update cannot capture. The corrector is trained on synthetic model--response pairs generated through the forward operator (Section~\ref{sec:learning_simulation}) and fine-tuned for real data using a composite objective combining data misfit, smoothness, and proximity to the EnsCGP reference (Section~\ref{sec:real_adaptation}).

\subsection{Ensemble-Conditional Gaussian Process (EnsCGP)}
\label{sec:EnsCGP}
The EnsCGP is the conditioning stage of the NPI workflow. From a prior ensemble of resistivity models and an observed MT response, it produces a physically admissible reference ensemble. It is the conditional Gaussian law obtained by treating the empirical moments of the prior ensemble and its forward-modeled responses as jointly Gaussian and conditioning on the observations under that approximation \cite{ravela_EnsembleconditionalGaussianProcesses_2026}. The single-pass update used here coincides with the Kalman analysis step while being interpreted as an ensemble approximation to Gaussian conditioning. The formulation follows ensemble-based Kalman methods \cite{evensen_EnsembleKalmanFilter_2003, Evensen_2009, ravela_2007, trautner_2020_informative_ensemble_Kalman_Learning, trautner_2020_informative_neural_ensemble_kalman_learning, saha_2024}: given a prior ensemble and its forward-modeled responses, EnsCGP computes a stabilized, locally linear update that adjusts the ensemble toward the observations.

\subsubsection{Formulation}
\label{sec:enscgp_formulation}
Let $N$ denote the number of ensemble members, $n_m$ the number of model parameters (depth layers in the 1D resistivity model), and $n_d$ the number of data components. Each MT response consists of apparent resistivity and phase at $n_f$ frequencies, so $n_d = 2n_f$. The prior ensemble is the matrix $\mathbf{A} \in \mathbb{R}^{n_m \times N}$, whose columns are individual members, and $\mathbf{D} \in \mathbb{R}^{n_d \times N}$ is the corresponding ensemble of forward-modeled responses. From these we form the empirical cross-covariance $\mathbf{C}_{AD}$ between model and data and the data covariance $\mathbf{C}_{DD}$; the matrix $\mathbf{R}$ encodes noise levels and the relative weighting of apparent resistivity and phase. Because the observed data $\mathbf{d}_{\mathrm{obs}} \in \mathbb{R}^{n_d}$ is a single realization while $\mathbf{D}$ contains $N$ members, we construct an observation ensemble $\mathbf{D}_{\mathrm{obs}} \in \mathbb{R}^{n_d \times N}$ by adding independent noise perturbations,
\begin{equation}
\label{eq:D_obs}
\begin{aligned}
\mathbf{D}_{\mathrm{obs}}
&=
\big[
\,\mathbf{d}_{\mathrm{obs}} + \boldsymbol{\epsilon}_{1},\;
\dots,\;
\mathbf{d}_{\mathrm{obs}} + \boldsymbol{\epsilon}_{N}
\,\big], \\
\boldsymbol{\epsilon}_{j} &\sim \mathcal{N}(\mathbf{0},\,\mathbf{R}).
\end{aligned}
\end{equation}
Perturbing the observations is a standard device for preserving ensemble spread under linear updates \cite{burgers_AnalysisSchemeEnsemble_1998}; the perturbations are independent draws and are not iteratively reused.
The EnsCGP update is then
\begin{equation}
\label{eq:kalman_update}
\mathbf{A}^{+} = \mathbf{A} + \mathbf{K}\left(\mathbf{D}_{\mathrm{obs}} - \mathbf{D}\right),
\qquad
\mathbf{K} = \mathbf{C}_{AD}\left(\mathbf{C}_{DD} + \mathbf{R}\right)^{-1},
\end{equation}
where $\mathbf{K}$ is the Kalman gain and $\mathbf{A}^{+}$ the updated ensemble. Since the correction is a linear combination of prior anomalies, updated members remain within the affine subspace spanned by the prior. The update cannot introduce structure absent from the prior. The cross-covariance $\mathbf{C}_{AD}$ couples model corrections to data discrepancies only along directions where model and data co-vary in the ensemble, so discrepancies orthogonal to those directions, including incoherent noise, do not drive the update, without any smoothing or filtering of the observed response.

\subsubsection{Stabilization}
\label{sec:enscgp_stabilization}
To improve numerical stability and suppress spurious correlations from finite ensemble sizes, we use a reduced-rank approximation of the sample covariance operator \cite{ravela_2007, ravela_RealtimeObservatoryLaboratory_2010}, controlling the dimensionality of the conditioning subspace. The observational covariance $\mathbf{R}$ may be deliberately inflated to stabilize the update when the ensemble is under-dispersive relative to observation--model discrepancies \cite{anderson_anderson_1999, anderson_2007, whitaker_hamill_2001, whitaker_hamill_2012}, and conditioning is localized to a nearest-neighbor subset of the prior ensemble in data space, limiting the influence of poorly matched samples. Localization and inflation are complementary: localization suppresses spurious long-range correlations from limited ensemble size, while inflation corrects under-dispersion relative to the model--data discrepancy.

\subsubsection{Distinction from iterative ensemble inversion}
\label{sec:enscgp_vs_eki}
EnsCGP is best understood as a representational object: the conditional Gaussian law rather than as an iterative solver. The update in Equation~\ref{eq:kalman_update} is one computational realization of this conditioning, applied once per observation. Although it shares the algebraic form of an ensemble Kalman analysis step, it is applied differently from iterative ensemble inversion methods such as ensemble Kalman inversion (EKI) and the ensemble smoother with multiple data assimilation (ES-MDA) \cite{Chen_2013, emerick_2013, iglesias_EnsembleKalmanMethods_2013}, which assimilate the same observation across multiple iterations for distinct reasons. EKI treats the ensemble as a derivative-free surrogate and is most naturally interpreted as an optimizer \cite{schillings_AnalysisEnsembleKalman_2017}, whereas ES-MDA distributes a single assimilation across several steps with correspondingly inflated covariances to better approximate the posterior for nonlinear forward operators \cite{emerick_2013}.

Iterating an ensemble update without a principled tempering or inflation schedule cannot be identified with Bayesian conditioning, since reusing the same observation injects no new information and drives the ensemble toward collapse. EnsCGP avoids this regime by construction, since the ensemble is not a gradient surrogate and no convergence or stopping criterion is involved. Any sequential structure in the NPI workflow reflects only changes in the localized reference ensemble used to build the empirical covariances, not repeated application of the same observation.

In EKI and ES-MDA, the cross-covariances used at iteration $k+1$ are computed from the ensemble iteration $k$ produced, so each update depends on the one before it and the same observation is reintroduced at every step. In EnsCGP, this dependency chain has length one, computed from the prior ensemble's own statistics and independent of any prior iteration.

The two iterative methods differ substantially once their guarantees are examined closely. With inflation coefficients satisfying $\sum_i 1/\alpha_i = 1$, ES-MDA applied to a linear Gaussian problem is shown in Emerick \& Reynolds (2013) \cite{emerick_2013} to be equivalent to a single Kalman update in both mean and covariance; its iterative structure serves to handle nonlinearity gradually, since a single step already recovers the correct posterior under the stated linear Gaussian conditions. The EKI guarantee, in the same setting, covers the mean only. Continued iteration collapses the ensemble toward a point, eliminating posterior spread, so retaining any spread requires stopping before that collapse occurs. EnsCGP reaches the correct mean and spread together, in the single step it takes.

\subsubsection{Output and propagation}
\label{sec:enscgp_output}
The EnsCGP outputs an ensemble of physically admissible resistivity models and corresponding forward-modeled MT responses, which serves as the reference input to the residual-learning stage and is propagated through subsequent stages.

\subsection{Residual-Learning Network (ResNet-1D)}
\label{sec:resnet}
The second stage refines the EnsCGP ensemble mean using a residual-learning convolutional neural network (ResNet-1D) \cite{He_2016_CVPR}. Consistent with the NPI design, the network does not perform an independent inversion. It is restricted to predicting structured residual corrections to an existing, physically admissible resistivity model, and never maps observations directly to resistivity. All data influence enters through the EnsCGP stage and through forward modeling, with data fit enforced explicitly by the forward operator during fine-tuning (Section~\ref{sec:real_adaptation}).

The central design decision is to learn residuals rather than full resistivity profiles, motivated by target-variance reduction. A direct inverse model must approximate the full nonlinear mapping from MT responses to resistivity, whose targets span the entire range of admissible structures and whose ill-posedness exposes the network to non-unique, noise-sensitive supervision. By contrast, the residual target
\begin{equation}
\label{eq:residual_target}
\mathbf{r}_{i}
=
\log \boldsymbol{\rho}^{\,\mathrm{true}}_{i}
-
\log \boldsymbol{\rho}^{\,\mathrm{EnsCGP,\,mean}}_{i}
\end{equation}
is the discrepancy from an estimate that already captures the dominant physically admissible structure, so its marginal variance is substantially smaller than that of $\log\boldsymbol{\rho}^{\,\mathrm{true}}_{i}$ and is localized in depth. This makes the supervised problem better-conditioned and limits the expressive role of the network.

The network input is the EnsCGP ensemble-mean, model-consistent MT response (log apparent resistivity and phase). Using forward-modeled responses rather than raw observations gives the corrector a physically consistent, noise-mitigated input. It is trained to approximate
\begin{equation}
\label{eq:residual_mapping}
f_{\theta}:\;
\mathbf{d}^{\,\mathrm{EnsCGP,\,mean}}_{i}
\longmapsto
\mathbf{r}_{i},
\end{equation}
and the refined resistivity estimate is recovered by adding the predicted residual to the EnsCGP ensemble mean in log space and exponentiating,
\begin{equation}
\label{eq:final_resistivity}
\hat{\boldsymbol{\rho}}^{\,\mathrm{NPI}}_{i}
=
\exp\!\left(
\hat{\mathbf{r}}_{i}
+
\log \boldsymbol{\rho}^{\,\mathrm{EnsCGP,\,mean}}_{i}
\right),
\end{equation}
so the neural output is a minimum-change adjustment to an existing physically admissible reference. The architecture is detailed in Appendix~A. We treat it as a baseline implementation rather than an optimized design, since the contribution is the modular framework, not a specific network topology.

Although trained on ensemble-mean quantities, the learned mapping $f_{\theta}$ is deterministic and applies to individual ensemble members. Each is propagated through the network using its own forward-modeled response, yielding a refined ensemble. The confinement to the prior span that the EnsCGP update guarantees does not carry over to this stage, since the residual correction is not a linear combination of prior anomalies (Section~\ref{sec:discussion_resnet}).

\subsection{Learning in Simulation}
\label{sec:learning_simulation}
\begin{figure}[!htbp]
\centering
\includegraphics[width=0.9\columnwidth]{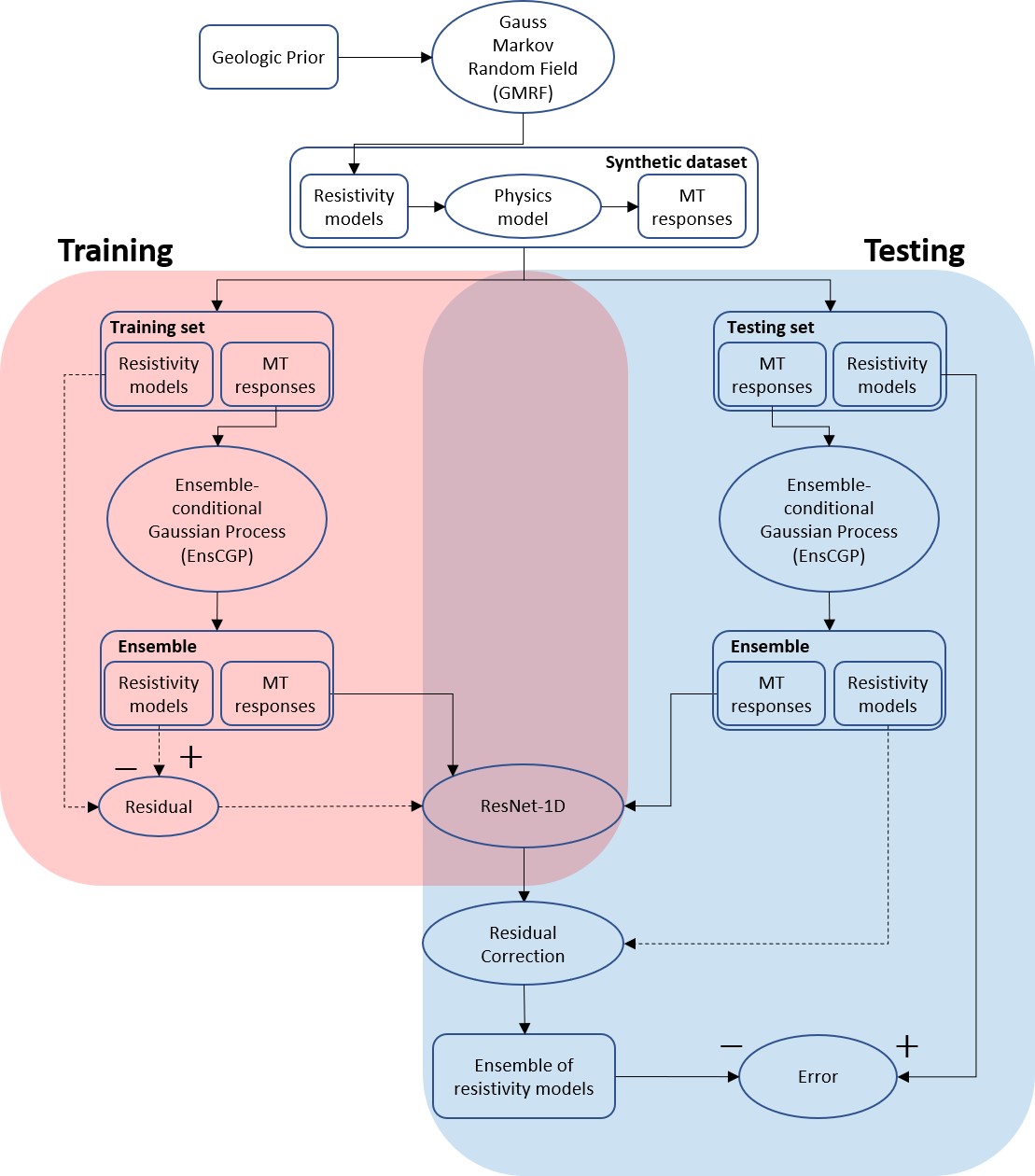}
\caption{
    NPI training and synthetic testing workflow. A geologic prior is projected into a low-rank GMRF, from which 1D resistivity profiles are sampled and forward-modeled to generate synthetic MT responses, split into training (red) and testing (blue) subsets. In training, noisy responses are conditioned through EnsCGP, and the ensemble-mean responses and the residuals between true and EnsCGP ensemble-mean log-resistivity profiles form the ResNet-1D inputs and targets. In testing, the trained ResNet-1D refines the EnsCGP ensemble-mean estimates for evaluation against the known truth.
}
\label{fig:training_workflow}
\end{figure}

Learning in simulation provides the controlled environment in which the residual network is constructed and trained (Figure~\ref{fig:training_workflow}). The aim is not to train a general inverse mapping but to expose the network to physically admissible examples of how EnsCGP first-guess estimates deviate systematically from known structure under realistic noise and prior variability. All inputs and targets are generated through forward modeling and ensemble conditioning, so the learning task stays anchored to MT physics and to the workflow itself.

\subsubsection{Prior Ensemble Construction}
\label{sec:prior_ensemble}
The NPI workflow requires a prior ensemble of forward-consistent, geologically plausible resistivity models but does not prescribe how it is built. Acceptable sources include layered random fields with prescribed correlation structure, borehole-constrained profiles, profiles from independent inversions of the same or nearby surveys, or profiles extracted from regional geological models. Broad, inflated priors are also admissible when site-specific information is unavailable. The prior defines the affine subspace within which EnsCGP conditioning operates. Since updated members remain within its span, the prior determines both what structures are reachable and how propagated variability is distributed across depth.

In the present work we use a Gauss-Markov random field (GMRF) constructed from a previous 3D resistivity model of the Gabbs Valley region \cite{peacock_2019_gv1, peacock_2021_gv2, peacock_siler_2021_grl}, a choice reflecting the availability of a regional model, not a requirement that NPI presuppose a 3D inversion. We extract an ensemble of 1D profiles, subtract the mean, and compute a singular value decomposition (SVD),
\begin{equation}
\label{eq:prior_svd}
\mathbf{M} - \bar{\mathbf{m}} = \mathbf{U}\,\mathbf{S}\,\mathbf{V}^{\top},
\end{equation}
where $\mathbf{M}$ contains the profiles and $\bar{\mathbf{m}}$ their mean. Retaining the leading $r$ components gives a low-rank representation of vertical variability, from which synthetic profiles are sampled,
\begin{equation}
\label{eq:gmrf}
\mathbf{m} = \bar{\mathbf{m}} + \mathbf{U}_{r}\,\mathbf{S}_{r}\,\mathbf{x},
\qquad
\mathbf{x} \sim \mathcal{N}(\mathbf{0},\sigma^{2}\mathbf{I}).
\end{equation}
We adopt $\sigma^{2}>1$ to inflate the prior covariance and explore a broader range of structures than the extracted profiles alone \cite{anderson_anderson_1999, anderson_2007, whitaker_hamill_2001, whitaker_hamill_2012}. The resulting prior concentrates variance in the depth intervals where the regional model varies most.

The dependence of ensemble spread on the prior is a substantive property of the workflow. Comparing the GMRF prior against a homogeneous half-space prior with depth-independent variance shows the depth structure of ensemble spread is inherited from the prior, not imposed by the workflow. This is examined in Section~\ref{sec:discussion_prior_dependence} and demonstrated in Appendix~C (Figure~\ref{fig:prior_sensitivity}).

\subsubsection{Forward Modeling and Frequency Grid}
\label{sec:forward_modeling}
Each profile is forward modeled with a 1D MT operator via Wait's recursion \cite{wait_1954}, computing apparent resistivity and phase on a fixed reference frequency grid spanning the field observational range. The fixed grid keeps the network input dimensionality consistent across training, testing, and application. Because the input to the ResNet-1D is the forward-modeled EnsCGP ensemble-mean response rather than the raw observation, it can always be evaluated on this grid regardless of a station's acquisition frequencies, handling frequency-configuration variability without retraining. To emulate field conditions, Gaussian noise is added at the complex-impedance level before computing apparent resistivity and phase. The resulting ensembles are then partitioned into training and testing subsets.

\subsubsection{Construction of Training Pairs}
\label{sec:training_pairs}
For each synthetic example, the noisy response is processed once through EnsCGP to produce a first-guess ensemble and its forward-modeled responses. The ensemble-mean response is the network input, and the target is the residual between the true and EnsCGP ensemble-mean log-resistivity profiles (Equation~\ref{eq:residual_target}), so the network learns only systematic corrections relative to the reference rather than an unconstrained inverse mapping.

\subsubsection{Training Procedure}
\label{sec:training}
Prior to training, the input channels (log apparent resistivity and phase) are standardized per frequency, and residual targets are standardized componentwise to stabilize optimization. The ResNet-1D is trained using mini-batch stochastic optimization with the AdamW optimizer \cite{loshchilov_adamw} and a Huber loss \cite{huber_1964}, which limits sensitivity to large residuals arising from high-contrast resistivity structures or unfavorable noise realizations. A \texttt{ReduceLROnPlateau} scheduler adapts the learning rate based on validation loss, and early stopping is employed to avoid overfitting. To reduce sensitivity to a particular train--validation split, training is performed using $K$-fold cross-validation, with normalization statistics retained for subsequent inference.

\subsubsection{Synthetic Testing}
\label{sec:testing}
Synthetic testing follows the training workflow with network parameters fixed. Each test response is conditioned through EnsCGP and refined by the trained network. Because the true profiles are known, these tests evaluate how the residual correction behaves across a range of subsurface structures, and under varying noise, before application to real data. Selected examples are shown in Figure~\ref{fig:synthetic_results_examples}.

\subsection{Adaptation to Real MT Data}
\label{sec:real_adaptation}
\begin{figure}[!htbp]
\centering
\includegraphics[width=0.7\columnwidth]{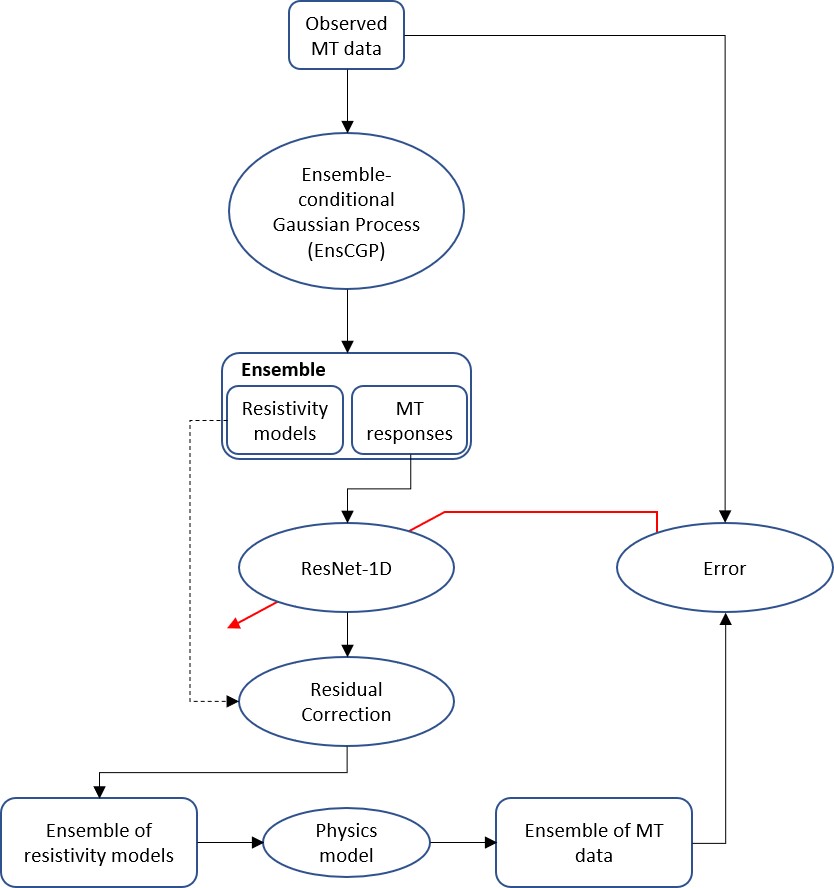}
\caption{
    Adaptation of the NPI to real MT data. From an observed MT response, EnsCGP produces an ensemble of physically admissible resistivity models and model-consistent responses. The ensemble-mean response is then passed through the pre-trained ResNet-1D to predict a depth-dependent residual correction. During fine-tuning, the network is updated under a physics-coupled objective (Equation~\ref{eq:realdata_loss}) enforcing minimum-change refinement around the EnsCGP estimate. At each optimization step, the network's resistivity estimate is forward modeled and its misfit against the observed response drives the parameter update. After convergence, the fine-tuned network is applied to each ensemble member, yielding a refined ensemble.
}
\label{fig:real_data_workflow}
\end{figure}

Applied to real observations, the NPI uses the same two-stage structure (Figure~\ref{fig:real_data_workflow}), but without true models for supervision. Adaptation instead relies on consistency with the observed data. This is achieved through a physics-coupled fine-tuning step in which the residual network is adjusted per station to improve forward-data fit under a regularized objective that limits departure from the EnsCGP reference.

Given an observed response $\mathbf{d}_{\mathrm{obs}}$ (log apparent resistivity and phase), EnsCGP first produces a stabilized ensemble of physically admissible models and forward-modeled responses from a localized nearest-neighbor subset of the prior. The ensemble means $\boldsymbol{\rho}^{\,\mathrm{EnsCGP,mean}}$ and $\mathbf{d}^{\,\mathrm{EnsCGP,mean}}$ form the reference state for refinement. Starting from the pre-trained ResNet-1D, station-wise fine-tuning takes $\mathbf{d}^{\,\mathrm{EnsCGP,mean}}$ as input, predicts a residual $\hat{\mathbf{r}}$, and forms $\hat{\boldsymbol{\rho}}$ via Equation~\ref{eq:final_resistivity}, minimizing
\begin{equation}
\label{eq:realdata_loss}
\mathcal{L}\!\left( \hat{\boldsymbol{\rho}} \right) =
\mathcal{L}_{\mathrm{phys}}
+
\lambda_{\mathrm{s}}\,\mathcal{L}_{\mathrm{s}}
+
\lambda_{\mathrm{r}}\,\mathcal{L}_{\mathrm{r}},
\end{equation}
where $\mathcal{L}_{\mathrm{phys}}$ enforces agreement with the observed data through explicit forward modeling, $\mathcal{L}_{\mathrm{s}}$ penalizes roughness, and $\mathcal{L}_{\mathrm{r}}$ constrains deviation from the EnsCGP ensemble mean.

This objective implements a minimum-change refinement principle: $\mathcal{L}_{\mathrm{r}}$ keeps stations already well explained by EnsCGP essentially unchanged, while $\mathcal{L}_{\mathrm{phys}}$ permits adjustments only where they reduce systematic misfit, and $\mathcal{L}_{\mathrm{s}}$ restricts those adjustments to structurally regular ones. Persistent misfit arises where the EnsCGP residual lies outside the prior span or where the forward operator is too nonlinear in the relevant directions for a locally linear update to correct, as well as from site-specific prior deviations, sparse frequency coverage, or limitations of the 1D representation. The network is therefore dormant when EnsCGP alone explains the data and active only where structured misfit warrants targeted correction.

\subsubsection{Physics-coupled Data Misfit Loss}
\label{sec:phys_loss}
Agreement with the observations is enforced by explicit forward modeling of the predicted model:
\begin{equation}
\label{eq:physics_loss}
\mathcal{L}_{\mathrm{phys}} =
\frac{1}{n_d}
\left\|
\mathbf{w} \odot \left( F\!\left( \hat{\boldsymbol{\rho}} \right) - \mathbf{d}_{\mathrm{obs}} \right)
\right\|_{2}^{2},
\end{equation}
where $F(\cdot)$ is the 1D MT forward operator and $\mathbf{w}$ holds inverse standard-deviation weights. Data fit is thus measured through explicit forward modeling rather than any learned surrogate.

\subsubsection{Smoothness Regularization}
\label{sec:smoothness}
Smoothness is enforced through first- and second-order derivative penalties on the resistivity profile,
\begin{equation}
\label{eq:smoothness_1}
\mathcal{L}_{\mathrm{s1}} =
\frac{1}{n_m - 1} \sum_{k=1}^{n_m - 1}
\left( \hat{\rho}_{k+1} - \hat{\rho}_{k} \right)^{2},
\end{equation}
\begin{equation}
\label{eq:smoothness_2}
\mathcal{L}_{\mathrm{s2}} =
\frac{1}{n_m - 2} \sum_{k=1}^{n_m - 2}
\left( \hat{\rho}_{k+2} - 2\hat{\rho}_{k+1} + \hat{\rho}_{k} \right)^{2},
\end{equation}
combined as
\begin{equation}
\label{eq:smoothness}
\mathcal{L}_{\mathrm{s}} =
(1-\alpha)\,\mathcal{L}_{\mathrm{s2}} + \alpha\,\mathcal{L}_{\mathrm{s1}},
\end{equation}
where $\alpha \in [0,1]$ controls the balance between curvature and first-derivative regularization.

\subsubsection{Reference-model Regularization}
\label{sec:smallness}
To ensure that neural refinement remains anchored to the EnsCGP reference, we include a reference-model penalty that constrains departures from the EnsCGP ensemble-mean resistivity model,
\begin{equation}
\label{eq:smallness}
\mathcal{L}_{\mathrm{r}} =
\frac{1}{n_m}
\left\|
\hat{\boldsymbol{\rho}} - \boldsymbol{\rho}^{\,\mathrm{EnsCGP,mean}}
\right\|_{2}^{2}.
\end{equation}
This term operationalizes the minimum-change refinement principle: corrections must be justified by reductions in $\mathcal{L}_{\mathrm{phys}}$ that exceed the cost of departing from the reference.

\subsubsection{Selection of the Target Misfit Level}
\label{sec:hyperparams}
The objective in Equation \ref{eq:realdata_loss} carries two weights, $\lambda_{\mathrm{s}}$, which suppresses oscillatory artifacts, and 
$\lambda_{\mathrm{r}}$, which enforces minimum-change refinement around the EnsCGP baseline, together with a third parameter $\alpha$ that enters through the smoothness penalty of Equation \ref{eq:smoothness} and balances curvature against first-derivative smoothing. These are included to give the objective flexibility, allowing the character of the penalty to be adapted to a survey or a parameterization. In the present application, $\lambda_{\mathrm{r}}$ and $\alpha$ are held fixed across all stations, while $\lambda_{\mathrm{s}}$ is adjusted automatically, following the same schedule at every station (Appendix~G).

In practice the selection reduces to a single choice: how closely the refined model is required to fit the data. Fine-tuning is halted by early stopping once the ensemble-mean nRMS reaches a target value, and this target governs how far the network is permitted to depart from the smooth EnsCGP reference in order to reduce $\mathcal{L}_{\mathrm{phys}}$. It therefore controls the same trade-off the weights encode, but through the stopping criterion rather than the loss weighting, and in a quantity that can be specified in advance from the estimated noise level rather than calibrated against the recovered profiles.

Demanding a lower misfit permits more fine-scale structure in the resistivity estimate, at the risk of fitting noise; tolerating a higher one keeps the estimate simpler and nearer the reference. The structure recovered by the workflow is thus indexed by the fit level requested of it, in the same sense that the target misfit governs model complexity in Occam-style regularized inversion \cite{constable_occam_1987}. For data with correctly specified Gaussian noise, the expected nRMS data misfit is 1, since the underlying misfit follows a chi-squared distribution with degrees of freedom equal to the number of data points. We adopt this value as the default target, so that structure appearing only under a tighter target can be identified as structure the data do not uniquely require. Appendix~E demonstrates this at two representative stations under two such targets.

The full set of fine-tuning hyperparameters and the protocol by which the target is set are reported in Appendix~G for reproducibility.

\subsubsection{Optimization and Ensemble Propagation}
\label{sec:finetuning_opt}
Fine-tuning uses the Adam optimizer with gradient clipping, a \texttt{ReduceLROnPlateau} scheduler, and early stopping on the physics-coupled misfit; batch-normalization layers are held in evaluation mode to preserve statistics from synthetic training. Although fine-tuning uses the ensemble-mean response, the network is deterministic and applies to individual members. Propagating the ensemble through it yields a refined ensemble.

\section{Results}
\label{sec:results}

\subsection{Synthetic Results}
\label{sec:synt_results}
We evaluate the NPI on a held-out synthetic test set independent of the training data. Each example is a 1D resistivity profile drawn from the GMRF prior (Section~\ref{sec:prior_ensemble}) with its forward-modeled MT response, contaminated with Gaussian noise; the dataset envelope is shown in Appendix~B (Figure~\ref{fig:synt_data_distribution}). The results below examine (i) the EnsCGP first-guess estimate, (ii) the bias reduction achieved through ResNet-1D refinement, and (iii) the degree to which ensemble spread is retained. Because EnsCGP and NPI share the same conditioning stage and prior, the comparison throughout isolates the contribution of the residual network as a controlled ablation. Differences between the two reflect the learned correction alone.

\subsubsection{Synthetic Data and Noise Model}
\label{sec:synt_data_noise}
Each test profile is drawn from the GMRF prior (Section~\ref{sec:prior_ensemble}) and forward modeled via Wait's recursion \cite{wait_1954}, which computes the complex MT impedance $Z$, from which apparent resistivity $\rho_a$ and phase $\phi$ are obtained \cite{weidelt_MT_transfer_function}. Noise is added at the impedance level with a 5\% relative magnitude error. Since $\rho_a \propto |Z|^2$, this propagates to first order as additive Gaussian noise of standard deviation 0.1 in $\log\rho_a$ and a phase standard deviation $\sigma_{\phi} \approx \tan^{-1}(0.05) \approx 2.86^{\circ}$:
\begin{equation}
\label{eq:rho_a_noise_model}
\begin{aligned}
\log \rho_{a}^{\mathrm{obs}} &= \log \rho_{a}^{\mathrm{true}} + \mathcal{N}(0,\,0.1^{2}), \\
\phi^{\mathrm{obs}} &= \phi^{\mathrm{true}} + \mathcal{N}(0,\,2.86^{2}).
\end{aligned}
\end{equation}

\subsubsection{Performance on Synthetic Test Examples}
\label{sec:synt_test}
We apply the full two-stage workflow to each test example. A single EnsCGP application produces a physically admissible first-guess ensemble, whose mean is refined by ResNet-1D. To quantify accuracy across the wide dynamic range of resistivity, we use the log-ratio misfit
\begin{equation}
\label{eq:log_ratio_misfit}
\Delta \log_{10} \rho(z) = \log_{10}\!\left( \frac{\rho_{\mathrm{est}}(z)}{\rho_{\mathrm{true}}(z)} \right),
\end{equation}
which symmetrically penalizes over- and under-estimation; zero is perfect recovery and $\pm 0.3$ is roughly a factor-of-two error.

\begin{figure}[!htbp]
    \centering
    \includegraphics[width=\columnwidth]{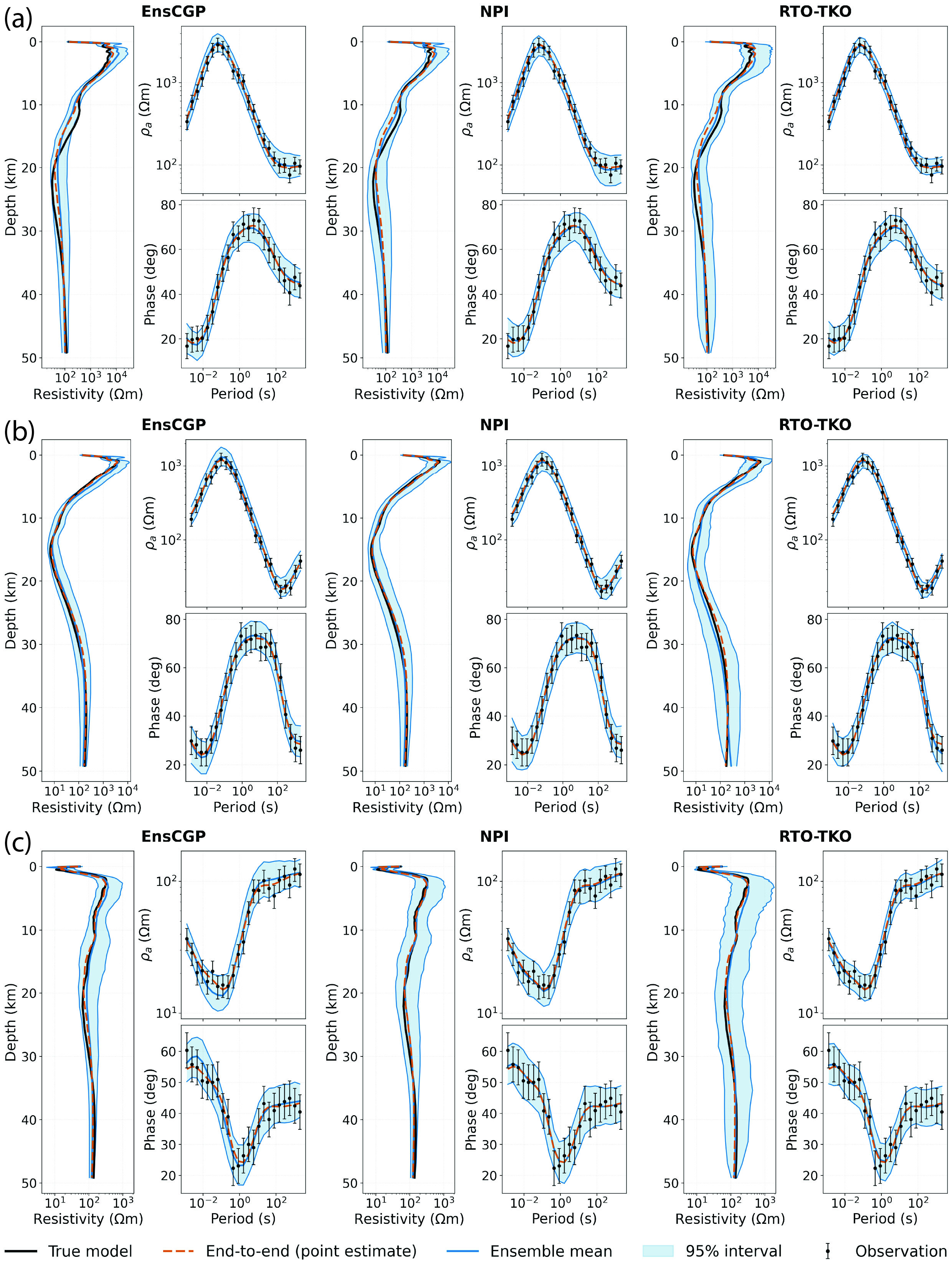}
    \caption{
        Inversion results for three representative synthetic test examples (a)–(c), compared across EnsCGP, NPI, and RTO-TKO (left to right within each row). For each method, the left panel shows the true resistivity model (black), ensemble mean (blue), and 95\% interval (shaded), together with the end-to-end point estimate (orange dashed). Right panels show the corresponding forward-modeled apparent resistivity and phase against the noise-contaminated observations (black, with error bars).
        }
    \label{fig:synthetic_results_examples}
\end{figure}

Figure \ref{fig:synthetic_results_examples} shows three representative examples, each compared across the EnsCGP and NPI stages of the proposed workflow and RTO-TKO \cite{blatter_part1_2022, blatter_part2_2022}, a randomize-then-optimize method that samples both the model and the regularization penalty weight, providing an estimate of uncertainty through repeated regularized inversion rather than ensemble conditioning. Each ensemble member is itself a deterministic, Occam-style regularized inversion, differing only in its sampled regularization weight and starting model, so comparison against RTO-TKO reflects Occam-style behavior at the level of individual members while additionally reporting the uncertainty context a single deterministic inversion could not provide. We adopt RTO-TKO as the baseline because it was developed and validated within EM geophysical inversion rather than adapted from an unrelated domain, runs on the same 1D forward operator without additional training or architecture reimplementation, and, estimating uncertainty through repeated regularized inversion rather than ensemble conditioning, provides a methodologically independent basis for judging EnsCGP and NPI on identical data.

The end-to-end point estimate (orange dashed) comes from a network trained to map the MT response directly to resistivity, rather than to a residual relative to the EnsCGP reference. It is overlaid on all three methods for reference, though it carries no ensemble of its own (Section~\ref{sec:discussion_resnet}). Across all three examples, the EnsCGP ensemble mean captures the dominant large-scale structure but shows amplitude bias and over-smoothing in some intervals. NPI reduces these biases while preserving that structure, closely tracking the end-to-end point estimate in the forward responses. The RTO-TKO ensemble mean is broadly comparable to EnsCGP and NPI across the three examples; its calibration relative to the other two methods is quantified below.

Table \ref{tab:synthetic_summary} summarizes performance for all four arms of the comparison over 5,000 test examples on a common 89-cell depth grid. Accuracy is the log-ratio misfit of Equation \ref{eq:log_ratio_misfit}, taken as a root mean square (RMS) over depth within each example and then averaged over examples. Data misfit is reported as nRMS, the RMS misfit normalized by the number of data points, computed from the response of the ensemble-mean model. 
Ensemble spread is summarized by a multiplicative \emph{factor of uncertainty} $U(z)$, the width of the central 95\% interval in $\log_{10}\rho$, which we interpret as a measure of spread under the assumed 1D prior and data representation. The half-width at depth $z$,
\begin{equation}
\Delta_{\log}(z) =
\tfrac{1}{2}
\left[
\log_{10}\rho_{97.5}(z) - \log_{10}\rho_{2.5}(z)
\right],
\label{eq:ci_log}
\end{equation}
where $\rho_{97.5}$ and $\rho_{2.5}$ are the corresponding percentiles in linear units, is converted to
\begin{equation}
U(z) = 10^{\Delta_{\log}(z)}.
\label{eq:factor_uncertainty}
\end{equation}
For a distribution approximately symmetric in $\log_{10}\rho$, the interval spans roughly $\rho_{50}/U$ to $\rho_{50}\,U$, so $U=2$ means the ensemble spans about half to twice the median resistivity; $U = 1$ corresponds to no spread, the ensemble collapsing to a single value. Table \ref{tab:synthetic_summary} reports $U$ averaged over depth and examples for each method.
The percentile column records where the true value falls within the ensemble, averaged over all examples and depths. A value of 50 indicates the truth is well-centered in the ensemble. A value below 50 means most ensemble members are more resistive than the truth, placing the truth toward the lower tail of the ensemble; a value above 50 means most ensemble members are less resistive than the truth, placing the truth toward the upper tail. The spread-to-error ratio compares the ensemble standard deviation against the RMS error of the ensemble mean, both in $\log_{10}$. A calibrated ensemble treats the truth as an additional member, so the truth's typical distance from the ensemble mean should match that of a member, and the ratio should be close to 1.0. 

The two stages behave as intended. EnsCGP alone reaches an RMS resistivity error of 0.131 between its ensemble-mean estimate and the true model. Unlike $U$, which describes the spread within a single example's ensemble, this measures how far the ensemble-mean estimate itself typically lands from the truth across different examples. An RMS error of 0.131 in $\log_{10}\rho$ means that for a true value of 100~$\Omega$m, the recovered estimate typically ranges from about 74 to 135~$\Omega$m across the test examples. The residual stage reduces this RMS resistivity error to 0.111, narrowing that range to about 78 to 129~$\Omega$m, and lowers the nRMS data misfit from 1.355 to 0.962.

The two ensembles miss calibration in different ways. EnsCGP has a spread nearly matched to its error, with a spread-to-error ratio of 0.97 that stays within 0.91 and 1.07 at every depth, and its coverage sits close to nominal at 88.6\% and 94.1\% against targets of 90\% and 95\%. The failure is in placement. The true value sits at the 42nd percentile of the ensemble on average, falling to the 35th near 3 km depth, meaning the truth sits off-center toward the ensemble's more-resistive tail at that depth. NPI reverses this. The truth is well-centered in the ensemble, falling between the 49th and 55th percentile at every depth. Its spread is nonetheless wider than its error requires, at a ratio of 1.40, so its coverage exceeds the target at both levels, reaching 96.8\% against 90\% and 98.6\% against 95\%.

These two departures from calibration have different consequences: an interval that is too wide sacrifices precision but still reliably contains the truth, while a truth that sits off-center in the ensemble is less likely to be captured by the interval at all, consistent with the EnsCGP coverage falling slightly below target.

The RTO-TKO coverage reaches 94.7\% (against 90\%) and 97.1\% (against 95\%), but this aggregate conflates two opposite failures occurring at different depths: the ensemble is too narrow for its errors at shallow depths and far too wide at depth. RTO-TKO also carries the widest intervals of the three ensembles, a mean $U$ of 2.62 against 1.85 for EnsCGP. It also has the largest RMS resistivity error between its ensemble-mean estimate and the true model.

\begin{table*}[t]
\centering
\caption{Performance on the Synthetic Test Set}
\label{tab:synthetic_summary}
\begin{tabular}{lccccccc}
\toprule
 & \multicolumn{2}{c}{Accuracy} & \multicolumn{5}{c}{Ensemble} \\
\cmidrule(lr){2-3}\cmidrule(lr){4-8}
Method & RMS & nRMS & mean $U$ & 90\% & 95\% & Percentile & Spread/error \\
\midrule
EnsCGP      & 0.131 & 1.355 & 1.85 & 88.6\% & 94.1\% & 42 & 0.97 \\
NPI         & 0.111 & 0.962 & 2.13 & 96.8\% & 98.6\% & 52 & 1.40 \\
RTO-TKO     & 0.135 & 1.011 & 2.62 & 94.7\% & 97.1\% & 48 & 1.42 \\
End-to-end  & 0.104 & 0.961 & --   & --     & --     & -- & --   \\
\bottomrule
\end{tabular}
\end{table*}

\subsection{Real-Data Results}
\label{sec:real_results}

\subsubsection{Overview}
\label{sec:real_data_overview}
We apply the NPI to MT data from the Gabbs Valley geothermal region of Nevada, USA \cite{peacock_2019_gv1, peacock_2021_gv2, peacock_siler_2021_grl}, which consists of 81 broadband stations across a region of Basin-and-Range extension with known geothermal activity. The same two-stage workflow validated on synthetic data is applied per station (Section~\ref{sec:real_adaptation}).

The aim is to demonstrate the framework's behavior under field conditions. The NPI results shown here derive solely from rotationally invariant 1D responses \cite{weaver_2000} at individual stations and cannot resolve 3D structure; comparison with the regional 3D MT inversion \cite{peacock_siler_2021_grl, peacock_2021_gv2} is a consistency check only, developed through the pseudo-2D cross-sections in Appendix~D.

\subsubsection{Study Area and MT Dataset}
\label{sec:study_area_data}
\begin{figure*}[!htbp]
    \centering
    \includegraphics[width=0.65\textwidth]{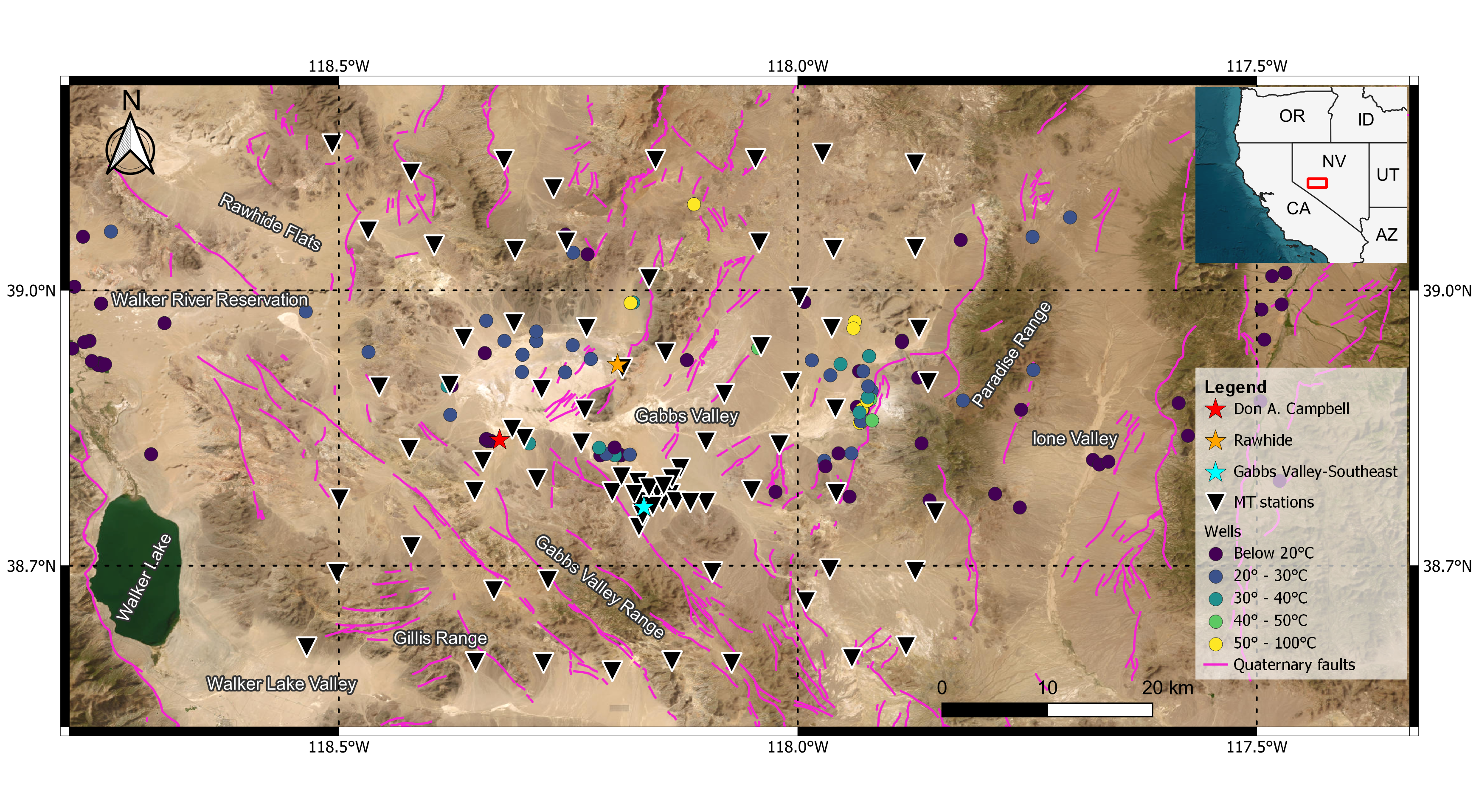}
    \caption{
    Gabbs Valley study area, Nevada, USA. MT stations (inverted black triangles) collected by the USGS over a satellite basemap with Quaternary fault traces (magenta) \cite{USGS_2006_Qfault}. Stars mark modern geothermal systems: Don A.\ Campbell (red), Gabbs Valley-Southeast (cyan), Rawhide (orange). Colored circles are wells binned by measured or geothermometry-derived temperature (legend). Inset: study-area location (red box) in the western United States.
    }
    \label{fig:gabbs_map}
\end{figure*}

Gabbs Valley, in west-central Nevada, lies at the structural transition between the dextral Walker Lane shear zone and the extensional Basin and Range province \cite{dilles_1995, bellier_1995, wesnousky_2005, lee_2020, faulds_2005, faulds_2008}. Miocene-to-recent normal and dextral faulting \cite{faulds_2008, Colgan_2006_Cenozoic, wernicke_1992} and elevated heat flow from crustal thinning \cite{lachenbruch_1978, blackwell_1983} create zones of enhanced fracturing and permeability that can drive deep fluid circulation and geothermal activity \cite{faulds_2008}. The region hosts several modern and fossil hydrothermal systems, including the Don A.\ Campbell \cite{Orenstein_2015}, southeastern Gabbs Valley \cite{craig_2021}, and Rawhide \cite{kratt_2008} geothermal systems. The study area, faults, geothermal features, and station locations are shown in Figure~\ref{fig:gabbs_map}.

The dataset comprises 81 broadband MT stations collected across southeastern Gabbs Valley by the U.S. Geological Survey (USGS) \cite{peacock_2019_gv1, peacock_2021_gv2}, each recording two horizontal electric and three magnetic field components. Data were acquired for $\sim$18~hours per station using alternating sampling rates (256~samples/s for $\sim$7.8~hours, 4096~samples/s for 10~minutes) and processed with the bounded-influence remote-reference algorithm (BIRRP) \cite{chave_2004_birrp}, yielding impedance tensors over $\sim$0.007--2048~s. As is typical for broadband MT, data quality falls toward the shortest periods (cultural and instrumental noise) and longest periods (reduced statistical robustness).

Because many stations show 2D or 3D behavior over portions of the band, indicated by phase-tensor diagnostics (Appendix~F, Figure~\ref{fig:phase_tensors}), the full impedance tensor is not interpretable within a strictly 1D framework. We therefore compute rotationally invariant impedances \cite{weaver_2000, szarka_1997, weaver_2006, marti_2009}, which collapse directional information into effective 1D responses. Although this discards anisotropy and higher-dimensional structure, it yields responses internally consistent with 1D assumptions. The resulting apparent resistivity and phase are the inputs to the real-data workflow.

\subsubsection{Application of NPI to Real MT Data}
\label{sec:NPI_real_data_application}
\begin{figure}[!htbp]
    \centering
    \includegraphics[width=\columnwidth]{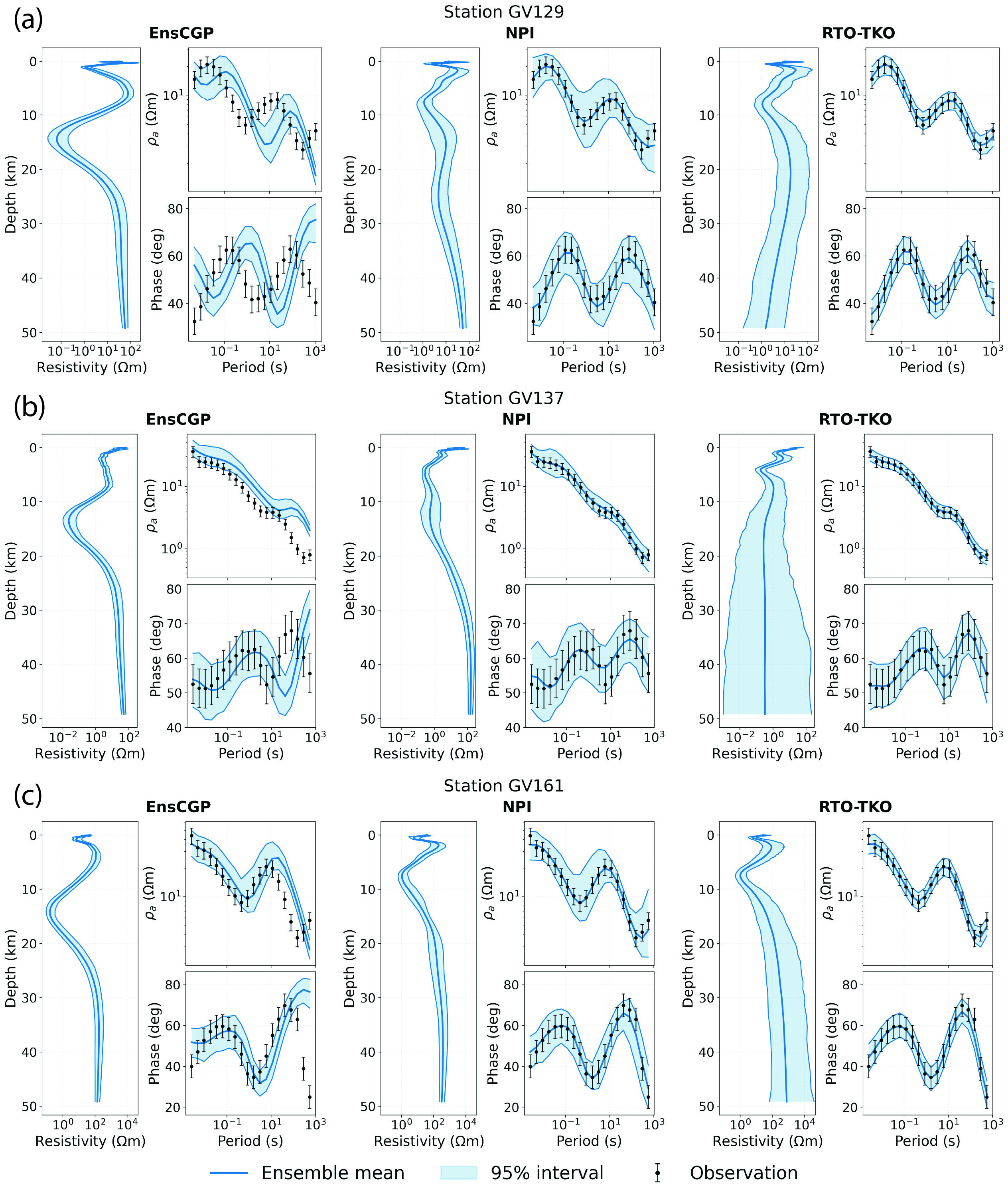}
    \caption{
        Station-level comparison of the EnsCGP, NPI, and RTO-TKO ensembles for stations
        GV129 (a), GV137 (b), and GV161 (c). For each method, the left panel shows the
        ensemble mean (blue) and its 95\% interval (shaded), and the right panels show
        the corresponding forward-modeled apparent resistivity and phase against the
        observed rotation-invariant data (black, with error bars).
        }
    \label{fig:real_npi_results_examples}
\end{figure}

Figure~\ref{fig:real_npi_results_examples} shows representative station-level behavior for three stations (GV129, GV137, GV161), compared across the EnsCGP, NPI, and RTO-TKO ensembles. EnsCGP provides a physically admissible ensemble broadly consistent with the observations, but a subset of stations shows persistent structured misfit. Station GV129 is the clearest case. The EnsCGP forward response diverges substantially from the data, while NPI brings it into close agreement. For such stations, the ResNet-1D supplies targeted corrections that reduce misfit while staying anchored to the EnsCGP estimate and without destabilizing the ensemble. Because these corrections need not remain within the prior span, they can fit structure the conditioning stage alone cannot.

RTO-TKO reaches a comparably close fit to NPI at the three stations. The two methods differ instead in model space, both in the ensemble spread and in the mean resistivity estimate. The RTO-TKO interval is consistently wider, by a factor of 1.8 to 3.0 in mean $U$ at GV129 and GV161. At GV137 the difference is far larger. At the deepest resolved layer, the RTO-TKO interval spans more than five orders of magnitude in resistivity, against less than a factor of two for NPI at the same depth. This divergence in interval width tends to coincide with divergence in the ensemble means. Above roughly 10~km, the ensemble means agree closely at all three stations. Below 10~km, the ensemble means diverge substantially at GV129 and GV137, where the RTO-TKO interval widens accordingly while the NPI interval does not. At GV161 the ensemble means remain in closer agreement at depth, yet the RTO-TKO interval still widens, indicating the width increase is not solely a consequence of the means diverging.

Two further practical considerations bear on choosing between these methods. NPI requires an upfront training cost, the synthetic pretraining of the ResNet-1D (Section~\ref{sec:learning_simulation}), that RTO-TKO does not. Training took approximately 1 hour on 40,000 synthetic samples under 5-fold cross-validation, using 8 GPU cores on WHOI's Hydra HPC cluster. This cost is incurred once and amortized across every subsequent station and survey using the same model parameterization; a change in the depth discretization or model dimensionality would instead require retraining (Section~\ref{sec:discussion_3D}). RTO-TKO draws are, in principle, near-independent and parallelizable \cite{blatter_part1_2022}; in the implementation used here, however, draws within a chain are serial, since each Gauss-Newton solve is warm-started and the smoothing weight follows a sequential Gibbs update (Appendix~G). At the per-station application stage, timed at station GV129 on an AMD Ryzen 5 PRO 4650U workstation (6 physical cores, 12 threads, no GPU; Appendix~G), EnsCGP conditioning and NPI fine-tuning and propagation together complete in 13.0~s for an ensemble of 500 members, while RTO-TKO requires 640.1~s for the same ensemble size, a factor of 49 slower.

\begin{figure}[!htbp]
    \centering
    \includegraphics[width=0.9\columnwidth]{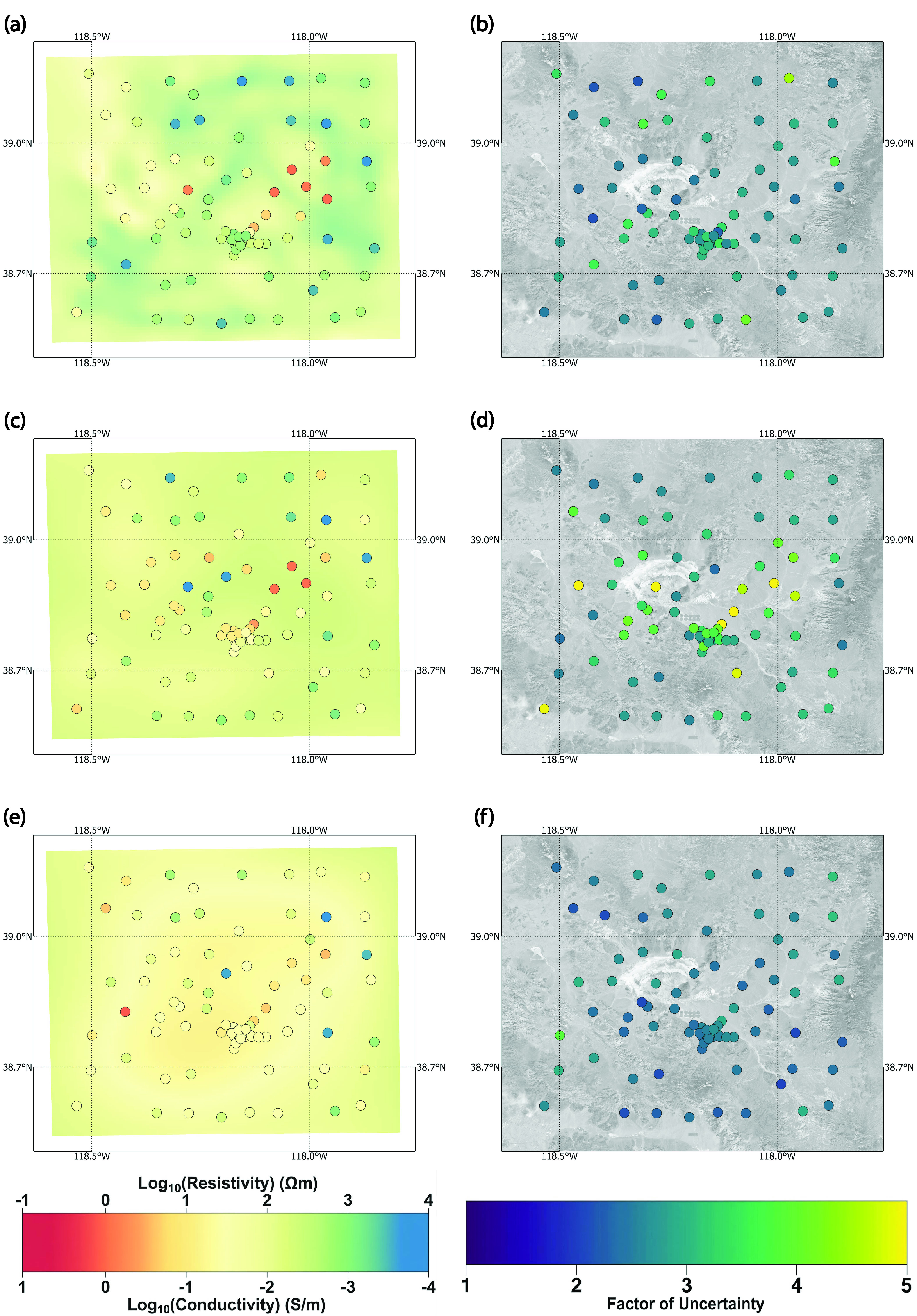}
    \caption{
    NPI ensemble-mean resistivity and spread across the Gabbs Valley survey. Panels (a), (c), (e): ensemble-mean estimates at $\sim$5, 15, and 25~km depth (colored circles) over depth slices of the regional 3D model \cite{peacock_siler_2021_grl, peacock_2021_gv2}, shown for context only. Panels (b), (d), (f): the multiplicative factor of uncertainty $U$ (Equation~\ref{eq:factor_uncertainty}), the width of the central 95\% ensemble interval in $\log_{10}\rho$, quantifying constraint within the assumed 1D model class. Larger $U$ indicates greater uncertainty. $U(z) = 2$, for example, means the 95\% interval spans from half to twice the ensemble median resistivity at that depth; $U(z) = 1$ corresponds to no spread.
    }
    \label{fig:real_depth_slices}
\end{figure}

Figure~\ref{fig:real_depth_slices} summarizes NPI ensemble-mean resistivity and spread across the survey at $\sim$5, 15, and 25~km depth. Panels (a), (c), (e) show the ensemble-mean estimates at each station over depth slices of the regional 3D model \cite{peacock_siler_2021_grl, peacock_2021_gv2} for context. Local discrepancies, expected where multi-dimensional structure or prior mismatch limit the 1D representation, are discussed in Section~\ref{sec:discussion}. The large-scale conductivity patterns are nonetheless broadly consistent with the 3D inversion (Appendix~D).

Ensemble spread across the survey is summarized by the factor of uncertainty $U(z)$ (Equation~\ref{eq:factor_uncertainty}). Panels (b), (d), (f) show $U$ at each depth, where smaller values indicate tighter constraint. Because $U$ inherits its structure from the prior ensemble, its depth-dependence is prior-dependent rather than intrinsic to the workflow; this dependency on the prior is examined further in Section~\ref{sec:discussion_prior_dependence} and Appendix~C.

\subsubsection{Data Misfit Diagnostics}
\label{sec:misfit_diagnostics}
\begin{figure}[!htbp]
\centering
\includegraphics[width=0.8\columnwidth]{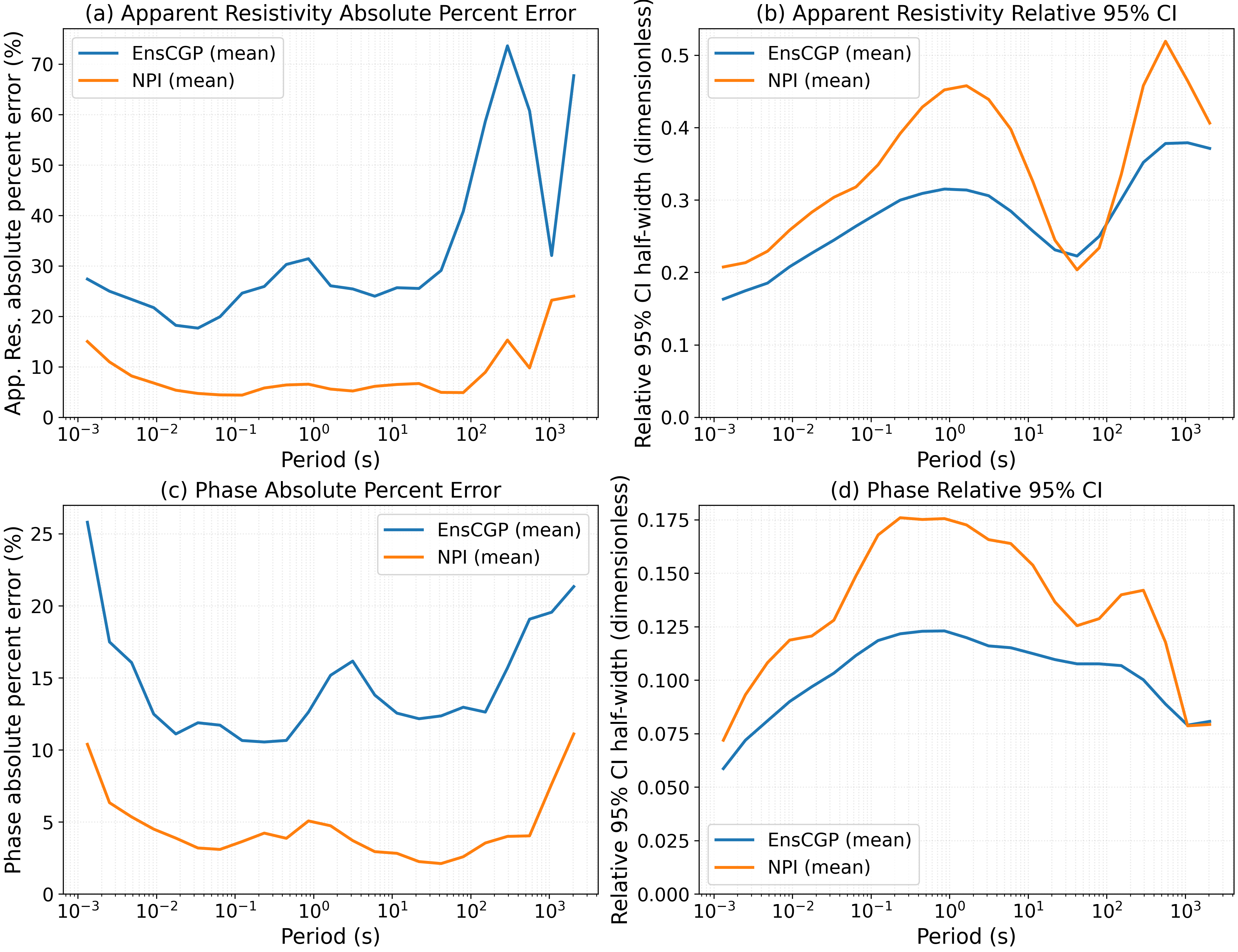}
\caption{
    Across-station data misfit and forward-response ensemble spread for the EnsCGP (blue) and NPI (orange) stages over all 81 stations. Panels (a), (c): mean absolute percent error between observed and ensemble-mean apparent resistivity and phase. Panels (b), (d): mean relative 95\% CI half-width (Equation~\ref{eq:rel_ci}). NPI reduces mid-band misfit in both apparent resistivity and phase relative to the EnsCGP baseline, while the relative ensemble spread remains comparable or modestly larger, showing the improved fit is not obtained by collapsing the ensemble.
}
\label{fig:real_data_misfits}
\end{figure}

Figure~\ref{fig:real_data_misfits} compares data misfit and forward-response ensemble spread for the EnsCGP and NPI stages across all 81 stations. Panels (a), (c) show the across-station mean absolute percent error between observed and ensemble-mean apparent resistivity and phase; panels (b), (d) show the across-station mean relative 95\% confidence-interval half-width,
\begin{equation}
\mathrm{RelCI}(f) =
\frac{\tfrac{1}{2}\big[x_{97.5}(f) - x_{2.5}(f)\big]}{x_{\mathrm{mean}}(f)},
\label{eq:rel_ci}
\end{equation}
where $x_{97.5}$, $x_{2.5}$, and $x_{\mathrm{mean}}$ are percentiles and mean of the forward-response ensemble at frequency $f$. For both stages, misfit is lowest over mid-periods, where signal-to-noise is highest, and rises toward the band edges where data quality degrades.

The EnsCGP baseline fits the observations to within roughly 25--30\% in apparent resistivity and 11--16\% in phase over the mid-band ($\sim$$10^{-1}$--$10^{1}$~s). NPI reduces both over the same band, bringing mean apparent-resistivity error below $\sim$10\% and mean phase error to a few percent. The relative ensemble spread under NPI is comparable to or modestly larger than under EnsCGP across the band. The slightly wider spread is expected, since the EnsCGP ensemble is in places too narrow to enclose the observations, whereas the NPI ensemble brackets them while fitting the mean more closely. The framework therefore tightens the fit to the observations while retaining the response variability that the factor of uncertainty summarizes.

\section{Discussion}
\label{sec:discussion}

\subsection{EnsCGP as a Representational Object}
\label{sec:discussion_enscgp}
As developed in Section~\ref{sec:enscgp_vs_eki}, EnsCGP is a single-pass, linear conditioning operator rather than an iterative inversion, avoiding the convergence-related failure modes of iterative ensemble methods; the stability seen in both the synthetic and real-data results follows.

Two consequences of this framing bear on interpretation. First, the quality of the EnsCGP estimate is fundamentally limited by the prior: because the update is linear, structures outside the prior span cannot be recovered, so where the prior is insufficient the workflow correctly produces persistent misfit rather than fabricating structure to fit the data. This is the failure mode the residual network is designed to address (Section~\ref{sec:discussion_resnet}).

Second, the EnsCGP spread admits a Bayesian reading. EnsCGP is the conditional Gaussian law obtained by treating the empirical moments of the prior ensemble and its forward-modeled responses as jointly Gaussian and conditioning exactly \cite{ravela_EnsembleconditionalGaussianProcesses_2026}; this joint Gaussian treatment is itself an approximation. Two restrictions qualify the reading. The first is the same prior span discussed above: the posterior mean shift lies in the range of the ensemble covariance, so a narrow prior characterizes uncertainty only within that subspace. The second is the Gaussian approximation, which broadening the prior does not remove, since the conditioning applied to a wider ensemble remains Gaussian and approaches the true posterior only insofar as that posterior is itself approximately Gaussian. These restrictions are what make the factor of uncertainty, discussed further in Section~\ref{sec:discussion_uncertainty}, a measure of constraint within the assumed model class rather than a posterior.

\subsection{Role of the Residual Network in a Modular Framework}
\label{sec:discussion_resnet}
The residual network addresses the prior-span failure mode: persistent structured misfit where the EnsCGP residual lies outside what a locally linear update can capture. This occurs when the prior is insufficient to span the relevant structure, or when the forward operator is too nonlinear in the relevant directions for a linear update to correct. Three design choices keep the network's role controlled: it learns low-variance residuals rather than full profiles, a choice motivated by tractability rather than accuracy; it takes the noise-mitigated EnsCGP mean response as input rather than raw observations; and it is regularized toward the EnsCGP reference under the composite objective (Equation~\ref{eq:realdata_loss}).

Table~\ref{tab:synthetic_summary} quantifies this trade-off. On resistivity accuracy, the end-to-end network reaches a slightly lower RMS resistivity error than the NPI ensemble mean, a modest and consistent edge across the test set. On nRMS data misfit, the two are effectively indistinguishable, with nearly identical values. Accuracy can be assessed only because the synthetic test set has a known true model; at a real station only nRMS data misfit is available, by which the two methods are equivalent.

The accuracy advantage of the end-to-end network in Table~\ref{tab:synthetic_summary} was measured using the ResNet-1D architecture of Appendix~A, a working baseline rather than an optimized design (Section~\ref{sec:resnet}), and does not on its own indicate how much either method's performance depends on that particular capacity. Appendix~H repeats the comparison with a backbone reduced 75-fold in parameter count, applied identically to the end-to-end mapping and to the NPI residual correction, and gives the end-to-end network an approximate uncertainty estimate through Monte Carlo dropout \cite{gal_2016_mc_dropout}, since it otherwise carries none.

With capacity matched, the accuracy edge of the end-to-end network disappears: NPI's RMS resistivity error and nRMS data misfit change only a few percent from the full-capacity values above, while the end-to-end network's roughly double, so its earlier edge rested on capacity the residual task never needed. NPI's advantage holds in distribution and widens under distribution shift, while its advantage over EnsCGP itself narrows under distribution shift. Calibration shows the same pattern: the end-to-end network's Monte Carlo dropout ensemble is markedly overconfident relative to NPI's in both settings. Full numbers are in Appendix~H (Table~\ref{tab:ablation}).

That is the uncertainty-aware property motivating the two-stage design (Section~\ref{sec:intro}): the ordering, EnsCGP first and ResNet-1D second, follows from where the ensemble originates, since the conditioning stage produces it and the residual network only refines its members rather than generating one itself.

The advantage of a residual target that is lower in variance and less tied to the structures of the training region is expected most directly in the per-station fine-tuning of Section~\ref{sec:real_adaptation} and the portability discussion of Section~\ref{sec:discussion_generalization}, settings where no true model is available and the network must adapt without labels. Table~\ref{tab:synthetic_summary} does not test that regime, since training and test profiles are drawn from the same synthetic prior. The synthetic distribution-shift ablation of Appendix~H is a step toward it, showing the residual target's accuracy advantage over the same-capacity end-to-end network widening rather than narrowing when the test prior departs from the training prior, though without the per-station fine-tuning or the real MT data that regime involves.

The end-to-end network is evaluated only on the synthetic test set and is not applied to the real Gabbs Valley stations, since doing so would require a per-station adaptation step. A scheme is possible in principle, for example anchoring the end-to-end network's fine-tuning to its own pretrained output rather than to a reference produced by EnsCGP, but constructing and validating one is outside the scope of the comparison presented here.

The essential contribution here is not the specific neural network design, which is a working baseline open to architecture search, structure learning, and uncertainty-calibrated variants. It is the modular separation of roles: EnsCGP produces a physically admissible reference within the prior span, the residual network corrects structured misfit the conditioning stage alone cannot, and both remain tied to the MT physics through forward modeling.

\subsection{Bias Removal and Generalization Across Settings}
\label{sec:discussion_generalization}
A consequence of the two-stage decomposition, beyond the ensemble-based measure of uncertainty, is that it reduces the dependence of the learned corrector on any particular geological setting. EnsCGP absorbs the systematic, prior- and location-dependent component of the model into a data-consistent baseline (Section~\ref{sec:resnet}), leaving the network to learn only the residual reflecting the generic limitations of a single linear update, namely structure outside the prior span and strong forward nonlinearity (Section~\ref{sec:discussion_resnet}), rather than the full amplitude and shape of the resistivity profile, which vary strongly from site to site.

A corrector trained on this residual target is therefore expected to transfer more readily across settings than one trained on a full resistivity profile. The synthetically trained network serves as a useful starting point wherever EnsCGP can supply a data-consistent baseline, and adaptation reduces to the lightweight per-station fine-tuning of Section~\ref{sec:real_adaptation} rather than retraining on a location-specific labeled distribution. In this sense, by removing prior-dependent bias from the training target, EnsCGP is what makes the residual-learning stage portable across geological settings. We frame this as a property the construction is designed to promote rather than one this study quantifies, since training and test profiles here are drawn from a common prior; systematic cross-region evaluation remains for future work.

\subsection{Prior-Dependence of Ensemble Spread}
\label{sec:discussion_prior_dependence}
The structure of the ensemble spread is a central interpretive question. Table \ref{tab:synthetic_summary} (Section~\ref{sec:synt_results}) quantifies this for the synthetic test set. The ensemble spread from EnsCGP is nearly correctly sized but the truth sits off-center within it, while for NPI the truth is well-centered but its spread exceeds its error, so calibration failure takes a different form at each stage rather than following a single pattern of inflation or collapse. In the real-data application, the factor of uncertainty $U(z)$ shows structured depth-dependence (Figure~\ref{fig:real_depth_slices}); this depth structure is inherited from the prior and should not be misattributed to an inherent characterization of the workflow.

Two factors set the propagated spread at each depth: the prior bounds the spread attainable by the conditioning stage, since the EnsCGP update confines all members to the prior span, and the data determine what fraction of that bound remains after conditioning. The residual correction then adjusts each member without that constraint, yet the depth structure inherited from the prior persists in the refined ensemble. Appendix~C shows this directly, with two priors producing nearly the same NPI point estimate but markedly different depth profiles of $U(z)$.

This prior-dependence is a substantive result that the workflow exposes by design. The factor of uncertainty should be read as a measure of constraint within the assumed model class under the chosen prior, which controls both what is reachable and how variability distributes across depth. Broad or homogeneous priors yield ensemble spreads reflecting data resolution alone, while informative regional priors yield correspondingly tighter spread where the prior is informative.

\subsection{Interpretation of the Factor of Uncertainty}
\label{sec:discussion_uncertainty}
Several outcomes call for careful reading: local discrepancies from the 3D model, elevated $U(z)$ at some stations, and residual misfit after refinement. These share common causes, including genuine multi-dimensional structure outside a 1D framework, prior misspecification, or training-distribution mismatch where the synthetic prior does not span the field structures, and are best read as diagnostics of the 1D representation's limits at specific stations rather than as framework failures. Elevated $U(z)$ specifically reflects non-uniqueness: small-scale resistivity oscillations the data do not uniquely require, a wide range of resistivities fitting the data equally well over a narrow interval, or a few weakly constrained members departing from the ensemble population. The fine-tuning target sensitivity test in Appendix~E supports this: detailed internal structure within conductive zones is not uniquely constrained by the data, consistent with the elevated spread there. Stations with elevated $U(z)$ therefore warrant examination for these same three causes, since $U(z)$ identifies where further analysis is needed without distinguishing among them.

The refined ensemble underlying $U(z)$ is not, in general, the conditional law of a stated model, so $U(z)$ should not be read as a calibrated Bayesian interval. The Bayesian reading available at the conditioning stage (Section~\ref{sec:discussion_enscgp}) does not extend to the residual correction: the corrector is fine-tuned per station against the same observation used in conditioning (Section~\ref{sec:real_adaptation}), so it is not a fixed map applied to the conditioned ensemble. This makes $U(z)$ a workflow-derived measure of ensemble spread, reflecting the combined effects of the prior, the conditioning step, the Gaussian approximation it rests on, and the learned, data-dependent correction.

The coverage test in Table~\ref{tab:synthetic_summary} measures empirical coverage directly, which is possible only because the synthetic test set carries a known true model. No comparable check exists in the real-data application, where the true resistivity structure is unavailable, so $U(z)$ at individual field stations carries no equivalent calibration guarantee. Even where ground truth exists, this test does not establish calibration against the true Bayesian posterior (Section~\ref{sec:discussion_limitations}). What the framework provides is an operational measure of uncertainty: $U(z)$ summarizes how tightly resistivity is constrained under the adopted workflow and prior, varying with data resolution and prior structure, and serves as a practical diagnostic of which recovered features are robust or weakly constrained.

\subsection{Toward Higher-Dimensional Applications}
\label{sec:discussion_3D}
The methodological contribution of this work is not specific to 1D. The EnsCGP update depends only on the empirical cross-covariance between model parameters and forward-modeled responses, so the same closed-form update applies to 2D sections or 3D volumes wherever the forward operator can be evaluated on each ensemble member. Four practical considerations bear on the extension:

\begin{itemize}
    \item \textbf{Prior construction.} The workflow is agnostic to how the prior ensemble is built: correlated random fields, geostatistical realizations, or regional models like the GMRF used here all serve as valid sources, with no change to the rest of the workflow.
    \item \textbf{Computational cost.} The forward operator in 3D is far more expensive than the 1D recursion, making the forward-ensemble cost dominant. Localization and reduced-rank conditioning become correspondingly more important. A learned or reduced-order surrogate for the forward operator could substitute for repeated exact forward evaluations across the ensemble, trading some accuracy for a substantial reduction in cost.
    \item \textbf{Retraining cost.} A change in depth discretization within a fixed dimension requires retraining only the network's final layer, since global average pooling makes the frequency features independent of grid resolution. A change in dimension, from 1D to 2D or 3D, changes the network's topology and is more substantial. The naive response, treating it as a new problem and retraining from scratch on a newly generated dataset, is dominated by the cost of forward simulation, modest in 1D ($\sim$1 hour; Section~\ref{sec:NPI_real_data_application}) but far larger at higher dimension. This is not necessary, however: initializing from a network trained at a lower dimension, so only a residual correction across dimension need be learned, or decoupling forward response resolution from the network's parameterization, could both reduce this cost substantially (Section~\ref{sec:discussion_limitations}). EnsCGP, needing no retraining, faces an analogous cost: its prior ensemble must be re-simulated at the new discretization, so both stages are ultimately driven by the cost of the forward operator.
    \item \textbf{Network architecture.} The residual network would adopt 2D or 3D convolutional or transformer layers, preserving the residual-learning principle while the physics-coupled objective generalizes.
\end{itemize}

The framework demonstrates several behaviors that arise from its modular design and carry over beyond the 1D setting, including stable ensemble conditioning, prior-informed uncertainty, targeted correction, and physics-coupled fine-tuning. We demonstrate it in 1D because the forward operator is inexpensive and the structure is transparent, but none of its components assume a 1D parameterization, positioning the framework to scale to higher-dimensional problems.

\subsection{Limitations and Outlook}
\label{sec:discussion_limitations}
Several limitations follow from the present scope. Rotation-invariant 1D responses discard the directional information in the full impedance tensor, so the results are effective 1D summaries of resistivity structure. As established in Section~\ref{sec:discussion_uncertainty}, $U(z)$ is not a calibrated Bayesian interval; what the framework reports is an ensemble spread whose reliability is established empirically, through the coverage test of Table~\ref{tab:synthetic_summary}, rather than by derivation. The fine-tuning weights are fixed across stations, and the target fit level is set from the estimated noise level rather than tuned per station (Section~\ref{sec:hyperparams}; Appendix~G); a formal joint sensitivity analysis over the weights themselves remains beyond the present scope. The pretrained ResNet-1D is tied to a fixed model parameterization, so adopting a different discretization or moving to 2D/3D requires retraining, whose cost is discussed in Section~\ref{sec:discussion_3D}; the architecture itself is a working baseline open to further optimization. We also do not separately isolate the contribution of synthetic pre-training from per-station fine-tuning; quantifying that decomposition is left to future work. Finally, the comparative validation here rests on a single alternative uncertainty estimator, RTO-TKO. Benchmarking against learned posterior methods that likewise trade upfront training cost for fast per-station inference would place the framework more completely among fast uncertainty-aware inversion schemes.

The comparisons in this study, evaluated throughout in the 1D setting, establish a necessary first step toward the credibility of the framework; NPI targets the higher-dimensional regime where Occam inversion and MCMC, though effective in 1D, become substantially more costly (Section~\ref{sec:intro}).

Beyond the extension to higher dimensions (Section~\ref{sec:discussion_3D}), several directions follow naturally. Alternative priors (e.g., borehole-constrained or structurally informed ensembles) substitute for the GMRF without changing the workflow. For surveys with substantially different frequency coverage, strategies such as frequency dropout during training \cite{srivastava_dropout_2014} or a learned input-projection layer could improve robustness to differing configurations. The retraining cost could be reduced further by generating training data progressively rather than all at once, or largely removed by a resolution-invariant architecture such as a neural operator; evaluating these alongside the strategies noted above is future work. Posterior-calibrated extensions of the factor of uncertainty, through conformal prediction or surrogate posterior sampling, would close the gap between workflow-derived spread and Bayesian uncertainty.

Finally, the two-stage design applies to any geophysical inverse problem with a well-defined forward operator and a means of generating model--response ensembles, where ensemble conditioning could supply a physically admissible baseline and a neural corrector refine the residual.

\section{Conclusions}
\label{sec:conclusions}
We introduced the Neuro-Physical Inverter (NPI), a modular two-stage framework coupling ensemble-based Gaussian conditioning (EnsCGP) with constrained residual correction (ResNet-1D), demonstrated in the 1D MT setting where the forward operator is inexpensive and the structure is transparent. The framework is uncertainty-aware by construction. An ensemble is conditioned, refined, and propagated through both stages, so every estimate carries an associated ensemble spread. The contribution is the framework and its empirical behavior, not the specific architecture, prior, or field application, and none of its components assume a 1D parameterization. The synthetic experiments establish two results: EnsCGP produces physically admissible reference ensembles through a single-pass conditioning step, and the residual network reduces ensemble-mean amplitude bias without destabilizing the ensemble. The application to 81 broadband stations in Gabbs Valley confirms this behavior under field conditions, reducing mid-band across-station mean error while retaining comparable ensemble spread.

The practical advantage of the ensemble formulation, discussed in Section~\ref{sec:discussion_uncertainty}, is that the inversion returns a resistivity model together with a depth-dependent, empirically validated measure of its reliability, an operational diagnostic flagging where the assumed model class and prior are well matched to the observations.

A key finding is that the depth-dependence of this ensemble spread is inherited from the prior rather than fixed by the workflow (Section~\ref{sec:discussion_prior_dependence}): the informed GMRF prior yields tighter spread at depth, while a homogeneous half-space prior yields spread that increases with depth, consistent with declining MT sensitivity, though robust near-surface features are recovered under both.

The NPI is not intended to replace established inversion methods such as Occam or trans-dimensional MCMC. What the framework offers instead is a scalable, uncertainty-aware inversion in which a physics-guided neural correction is integrated with ensemble conditioning, verified here in a controlled 1D setting. Because the EnsCGP update and the residual-learning principle are dimension-agnostic, the same design is positioned to scale to 2D and 3D, where an uncertainty-aware framework is most valuable and full posterior sampling is most costly.

\section*{Acknowledgments}
Codes used in this study are openly available on GitHub and archived on Zenodo \cite{kim_npi_mt_2026}. The MT dataset from the Gabbs Valley region was collected by the USGS and can be accessed via Peacock (2019) \cite{peacock_2019_gv1} and Peacock et al. (2021) \cite{peacock_2021_gv2}. The authors have no conflicts of interest to disclose. JDK was supported by internal graduate fellowships provided by the Woods Hole Oceanographic Institution (WHOI). This material is based in part upon work supported by the Bangkok Bank, Schmidt Sciences, Community Jameel for Jameel Observatory CREWSnet, and by the MIT Climate Grand Challenges. 

{
\appendices

\section*{Appendix A: ResNet-1D Architecture}
\label{app:resnet_architecture}
The ResNet-1D is a 1D residual-learning convolutional network that predicts depth-dependent corrections to the EnsCGP ensemble-mean resistivity profile, treated here as a working baseline rather than an optimized design.

The two input channels are log apparent resistivity and phase as functions of frequency, both forward modeled from the EnsCGP ensemble-mean profile rather than taken from raw observations. An initial 1D convolutional stem projects these into a 32-channel representation, followed by four residual blocks that expand and contract the channel dimension (32$\rightarrow$64$\rightarrow$128$\rightarrow$64$\rightarrow$32) to extract multi-scale frequency-domain features. Each block comprises two 1D convolutional layers with batch normalization and ReLU activation, a dropout layer, and a skip connection. Global average pooling then aggregates across the frequency band into a fixed-length feature vector independent of the number of input frequencies, and a fully connected module maps it to a depth-dependent residual correction in log-resistivity space. The correction is added to the EnsCGP ensemble-mean log-resistivity profile and exponentiated to produce the refined NPI estimate (Equation~\ref{eq:final_resistivity}). Training details are in Section~\ref{sec:training}.

\begin{figure*}[!htbp]
\centering
\includegraphics[width=\textwidth]{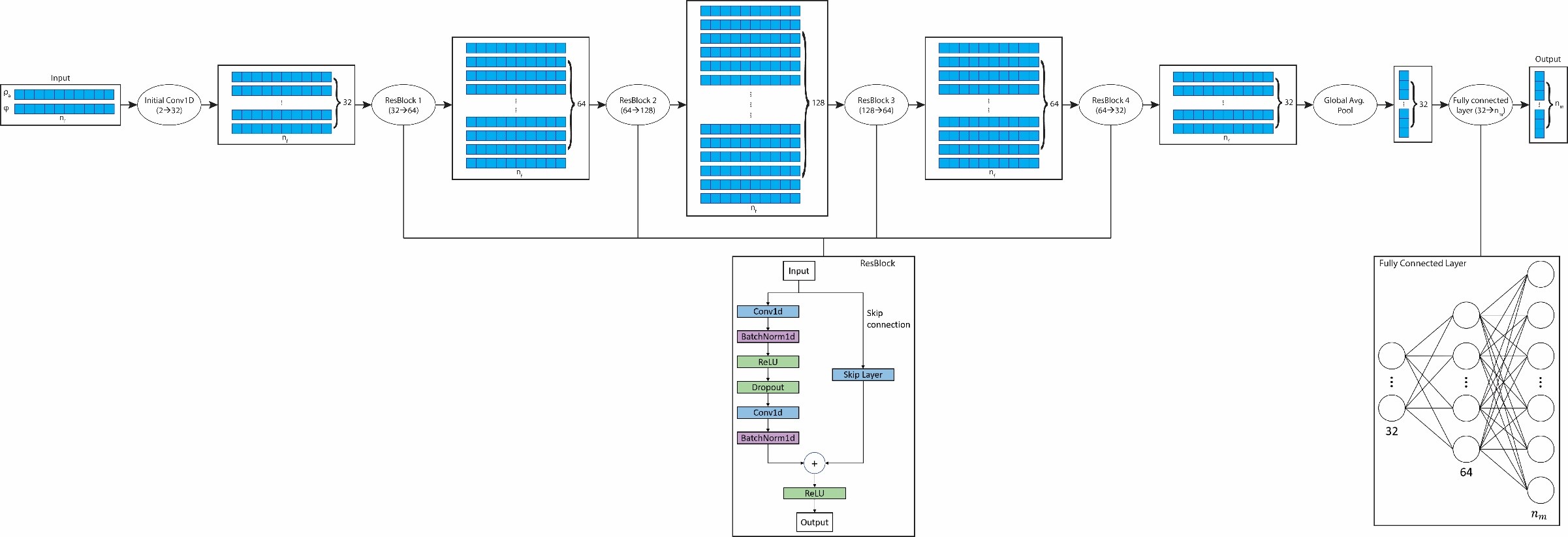}
\caption{
    Architecture of the ResNet-1D. Two input channels (log apparent resistivity and phase) pass through a 1D convolutional stem, four residual blocks expanding and contracting the channel dimension (32$\rightarrow$64$\rightarrow$128$\rightarrow$64$\rightarrow$32), global average pooling, and a fully connected output module that produces a depth-dependent residual correction in log-resistivity space, added to the EnsCGP ensemble mean to give the refined NPI estimate. Insets show a residual block and the fully connected module.
}
\label{fig:resnet_architecture}
\end{figure*}

\section*{Appendix B: Synthetic Dataset Distribution}
\label{app:synt_data}
Figure~\ref{fig:synt_data_distribution} summarizes the synthetic dataset, illustrating the distribution of 1D resistivity models sampled from the GMRF prior and the range of clean MT responses they imply. The three representative test examples used throughout this work, shown with full inversion results across EnsCGP, NPI, and RTO-TKO in Figure~\ref{fig:synthetic_results_examples} of the main text, are drawn from this same distribution. 

\begin{figure}[!htbp]
\centering
\includegraphics[width=0.8\columnwidth]{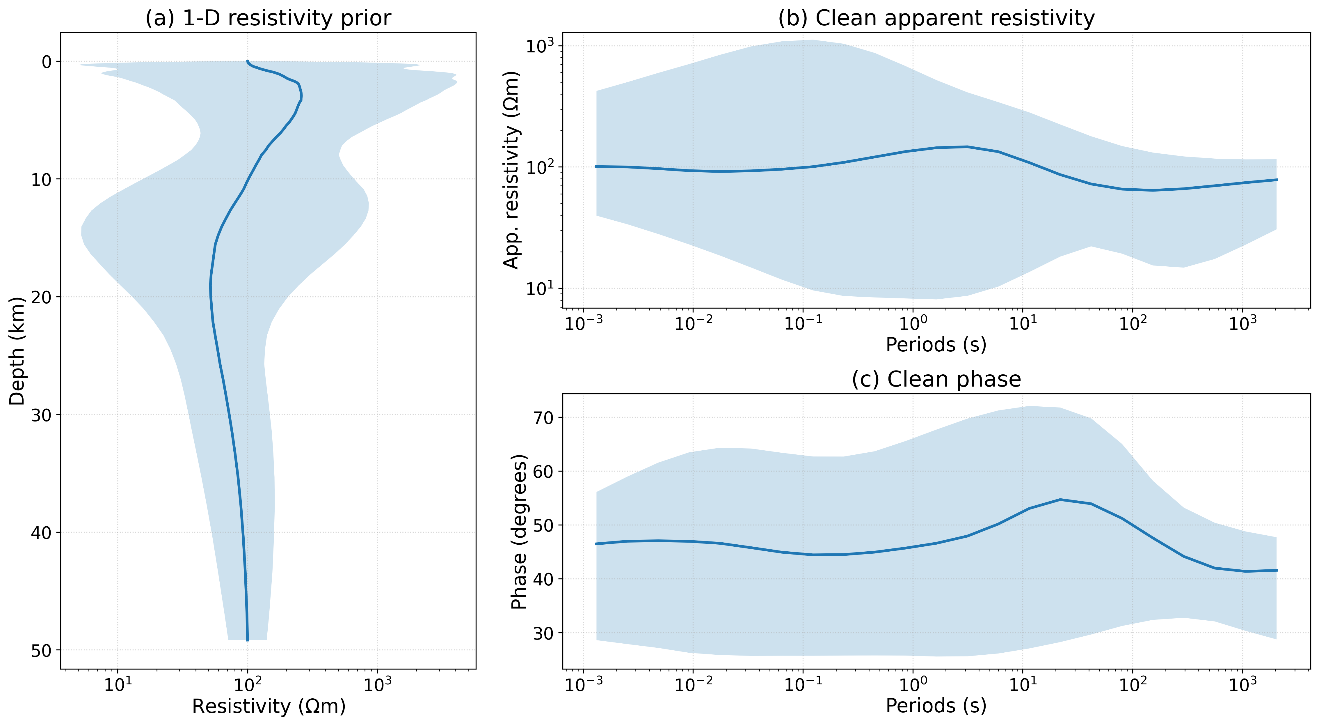}
\caption{
    Synthetic dataset for training and evaluation. (a) Distribution of 1D resistivity models sampled from the reduced-rank GMRF prior; solid curve is the ensemble median, shaded region the 5--95\% interval. (b--c) Corresponding clean forward-modeled apparent resistivity and phase responses, same median and percentile convention. The spread reflects geologic variability encoded in the prior.
}
\label{fig:synt_data_distribution}
\end{figure}

\section*{Appendix C: Prior Sensitivity Comparison}
\label{app:prior_sensitivity}
To demonstrate that the depth structure of the factor of uncertainty $U(z)$ is inherited from the prior rather than fixed by the workflow, we compare NPI results at station GV09 under two priors, namely the regionally informed GMRF prior constructed from the Gabbs Valley 3D model, and a homogeneous half-space prior with constant mean resistivity and depth-independent variance, encoding no particular assumptions about how resistivity varies with depth.

Figure~\ref{fig:prior_sensitivity}(a) shows results under the informed GMRF prior. The prior ensemble concentrates variance in the shallow section where the regional reference model varies most, and after conditioning and refinement, the NPI ensemble mean reproduces the observed responses with spread that stays close to constant with depth. Figure~\ref{fig:prior_sensitivity}(b) shows results under the homogeneous prior. The prior ensemble spans a wide resistivity range uniformly with depth, and the refined NPI ensemble spread instead widens with depth, consistent with the decrease in MT sensitivity at depth.
The NPI ensemble-mean resistivity profile itself changes little between the two priors. Averaged across the full depth profile, the two mean profiles differ by a factor of about 1.2 typically, and by no more than a factor of 1.7 at any single depth. The corresponding forward responses are similarly close, differing on average by 7\% in apparent resistivity and 3\% in phase. Starting from two very different priors therefore leads to nearly the same point estimate, in both model space and data space.

Fit to the observations is comparably good under both priors. Fine-tuning at GV09 halts once the ensemble-mean nRMS misfit reaches the target level of 1.0 (Section~\ref{sec:hyperparams}). The achieved misfit is 0.56 under the GMRF prior and 0.78 under the homogeneous prior, both below this target. An nRMS below the nominal target is not a concern for an ensemble-based method as it is for a deterministic inversion: the nRMS here is computed from the ensemble mean, so a lower value means the mean sits closer to the observation, and thus that the observation is well-centered within the resulting ensemble. Averaged over the frequency band, apparent-resistivity error against the observations is 5\% under the GMRF prior and 9\% under the homogeneous prior, and phase error is 2\% and 3\% respectively. The GMRF prior reaches a tighter fit, but both priors satisfy the fine-tuning stopping criterion and reproduce the observed responses to a comparable standard.

What differs sharply is the depth structure of the ensemble spread. Splitting the profile at 10~km, mean $U$ under the GMRF prior is 2.57 above 10~km and 2.79 below, a deep-to-shallow ratio of 1.09, essentially flat with depth. Under the homogeneous prior, mean $U$ rises from 1.69 above 10~km to 3.74 below, a ratio of 2.22, more than doubling. Together these indicate that the depth-dependence of $U(z)$ is set by the prior rather than by the workflow, while the point estimate remains comparatively insensitive to which prior is used, so the workflow produces meaningful results without a regionally informed prior and correspondingly tighter, better-constrained uncertainty where one is available.

\begin{figure}[!htbp]
\centering
\includegraphics[width=0.9\columnwidth]{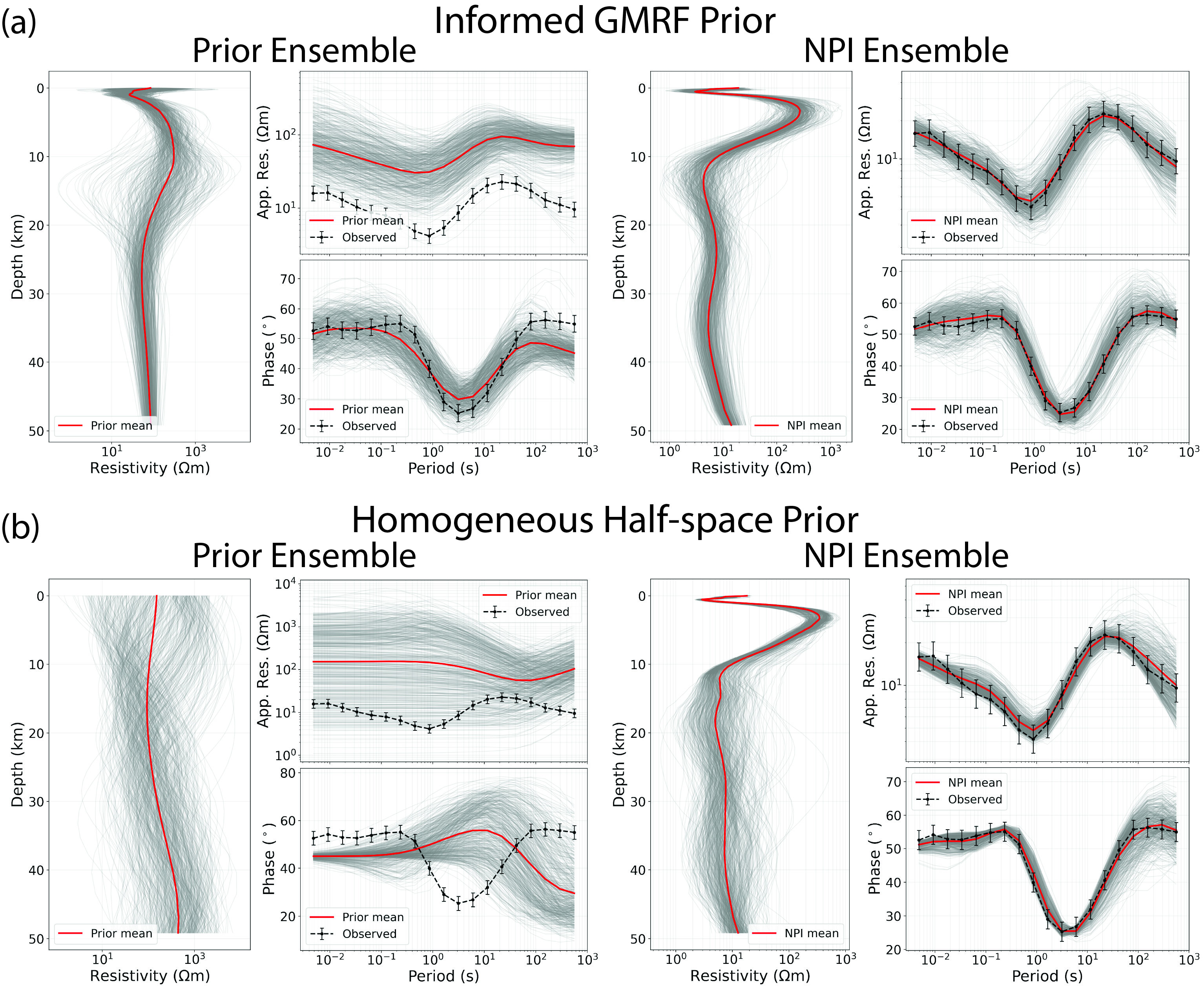}
\caption{
    Prior sensitivity comparison at station GV09. (a) Regionally informed GMRF prior. (b) Homogeneous half-space prior with constant mean resistivity and depth-independent variance. In each row, the left pair of panels shows the prior resistivity ensemble (gray) with prior mean (red) and the corresponding forward-modeled apparent resistivity and phase against the observed rotation-invariant MT data (black dashed with error bars); the right pair shows the NPI-refined resistivity ensemble (gray) with NPI ensemble mean (red) and the corresponding forward responses against the observations. The NPI ensemble mean reproduces the observed responses under both priors, but ensemble spread behaves differently with depth in each case, discussed in the text.
}
\label{fig:prior_sensitivity}
\end{figure}

\section*{Appendix D: Pseudo-2D Cross-Sections and Consistency with Regional 3D Inversion}
\label{app:pseudo-2d}
To place the station-wise NPI results in spatial context, we stitch the NPI ensemble-mean 1D estimates at stations projected onto two profiles, A-A' and B-B', within the Gabbs Valley survey area (Figure~\ref{fig:pseudo-2d}), limited to the upper 5~km where near-surface structure is best resolved. As noted in Section~\ref{sec:real_data_overview}, this is a qualitative consistency check rather than a geological interpretation superseding the 3D inversion, since the NPI results derive solely from rotation-invariant 1D responses and cannot resolve lateral continuity or anisotropy.

The most prominent feature in the NPI sections is a shallow conductive layer at roughly 0--2~km, detected consistently across stations along both profiles and coinciding with the conductive near-surface structure in the 3D inversion. Its depth range is consistent with a geothermal clay cap, a hydrothermal-alteration feature widely observed in geothermal fields \cite{newman_2008, Cumming2009GEOTHERMALRC, munoz_ExploringGeothermalResources_2014, wannamaker_2019, craig_2021}. The NPI workflow does not constrain the layer's thickness or internal structure, but its consistent detection suggests it is a first-order characteristic of the near-surface electrical structure. Below it, both the 3D and NPI sections show an increase in resistivity within the upper crust. Interpretation beyond 5~km requires care given the 1D invariant framework and is outside the present scope.

The factor-of-uncertainty panels (e, f) show relatively low $U(z)$ through the near-surface conductor at most stations, indicating it is comparatively well-constrained, with spread increasing toward the base of the interval as MT sensitivity decreases. Locally elevated $U(z)$ at near-surface depths at some stations may reflect site-specific effects, prior misspecification, or multidimensional structure.

Overall, the NPI pseudo-2D sections reproduce the large-scale near-surface conductivity patterns of the regional 3D inversion along both profiles, despite relying only on rotation-invariant 1D responses, supporting the framework as a practical tool for recovering regional electrical structure with an ensemble-based characterization of constraint.

\begin{figure}[!htbp]
\centering
\includegraphics[width=0.9\columnwidth]{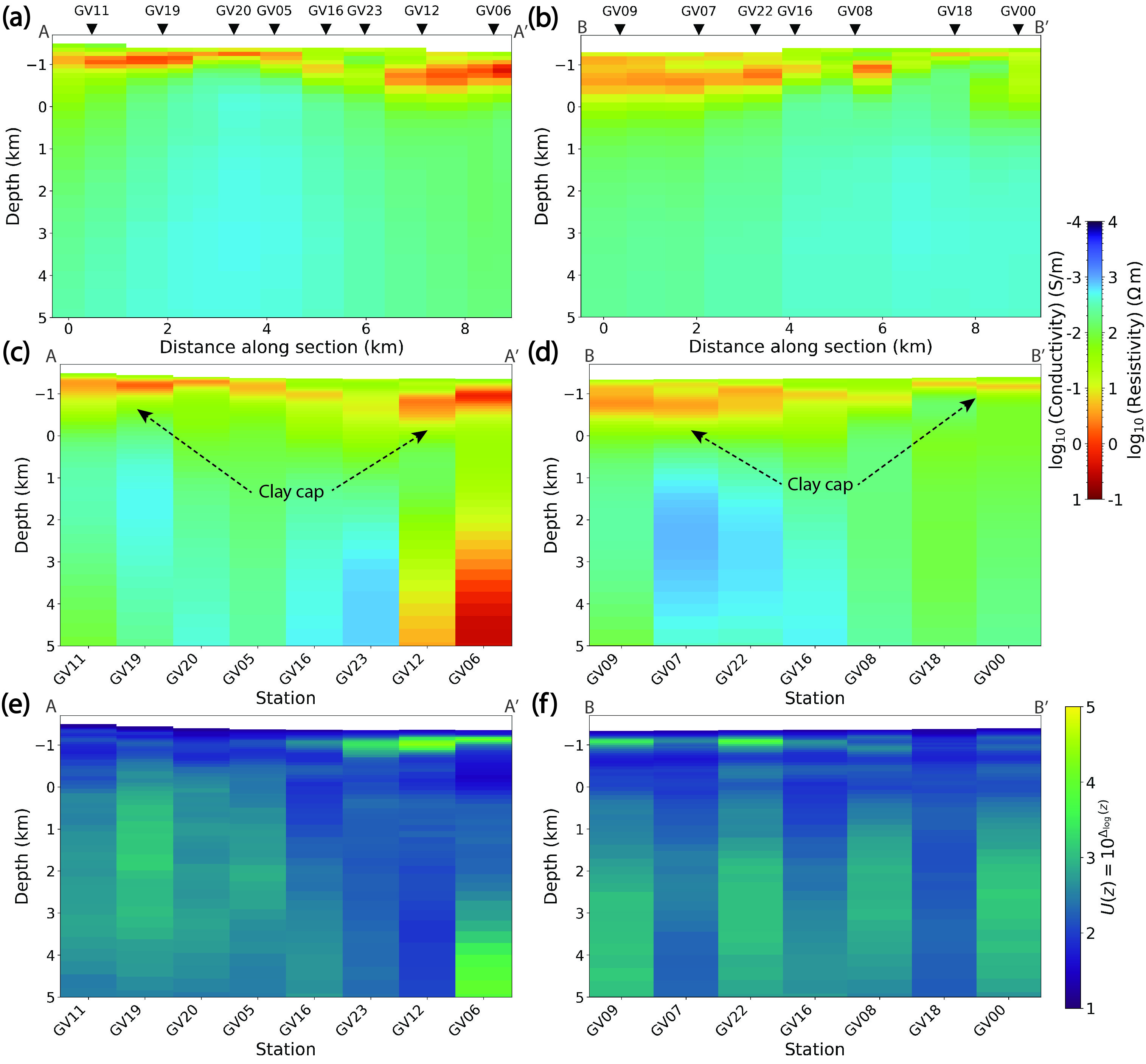}
\caption{
    Regional 3D MT inversion cross-sections \cite{peacock_2021_gv2} compared with pseudo-2D sections from the NPI 1D inversions along profiles A-A' (left) and B-B', upper 5~km. Profiles A-A' and B-B' run approximately southwest--northeast and northwest--southeast across the Gabbs Valley survey area. (a, b) resistivity from the 3D inversion. (c, d) NPI pseudo-2D sections, stitched from ensemble-mean 1D estimates at stations projected onto each profile. (e, f) factor of uncertainty $U(z)$ from the NPI ensembles. Station locations are marked at the top of each panel. A shallow conductive layer consistent with a geothermal clay cap is indicated in (c, d).
}
\label{fig:pseudo-2d}
\end{figure}

\section*{Appendix E: Sensitivity of NPI Results to the Fine-Tuning Target}
\label{app:regularization_test}
Station-wise fine-tuning (Section~\ref{sec:real_adaptation}) is driven by a target data misfit level rather than by direct selection of the weights $\lambda_s$, $\lambda_r$, $\alpha$ (Section~\ref{sec:hyperparams}). To show how the target governs the recovered structure in practice, we repeat fine-tuning at two representative stations (GV06 and GV120) under a target nRMS of 1.00, the nominal target corresponding to the expected data misfit level, and a representative loosened value of 1.50 (Figure~\ref{fig:nrms_target_sensitivity}).

Relaxing the target from 1.00 to 1.50 suppresses finer-scale structure within the conductive zone at both stations, producing a comparatively simpler resistivity profile, while the large-scale conductive feature persists under either target. Structure that appears only under the tighter target is therefore structure the data do not uniquely require, consistent with the non-uniqueness argument of Section~\ref{sec:discussion_uncertainty}. The fit level acts as the operative control on model complexity in the workflow, with the fixed weights setting the character of the regularization and the target setting how strongly it binds.

\begin{figure}[!htbp]
\centering
\includegraphics[width=0.7\columnwidth]{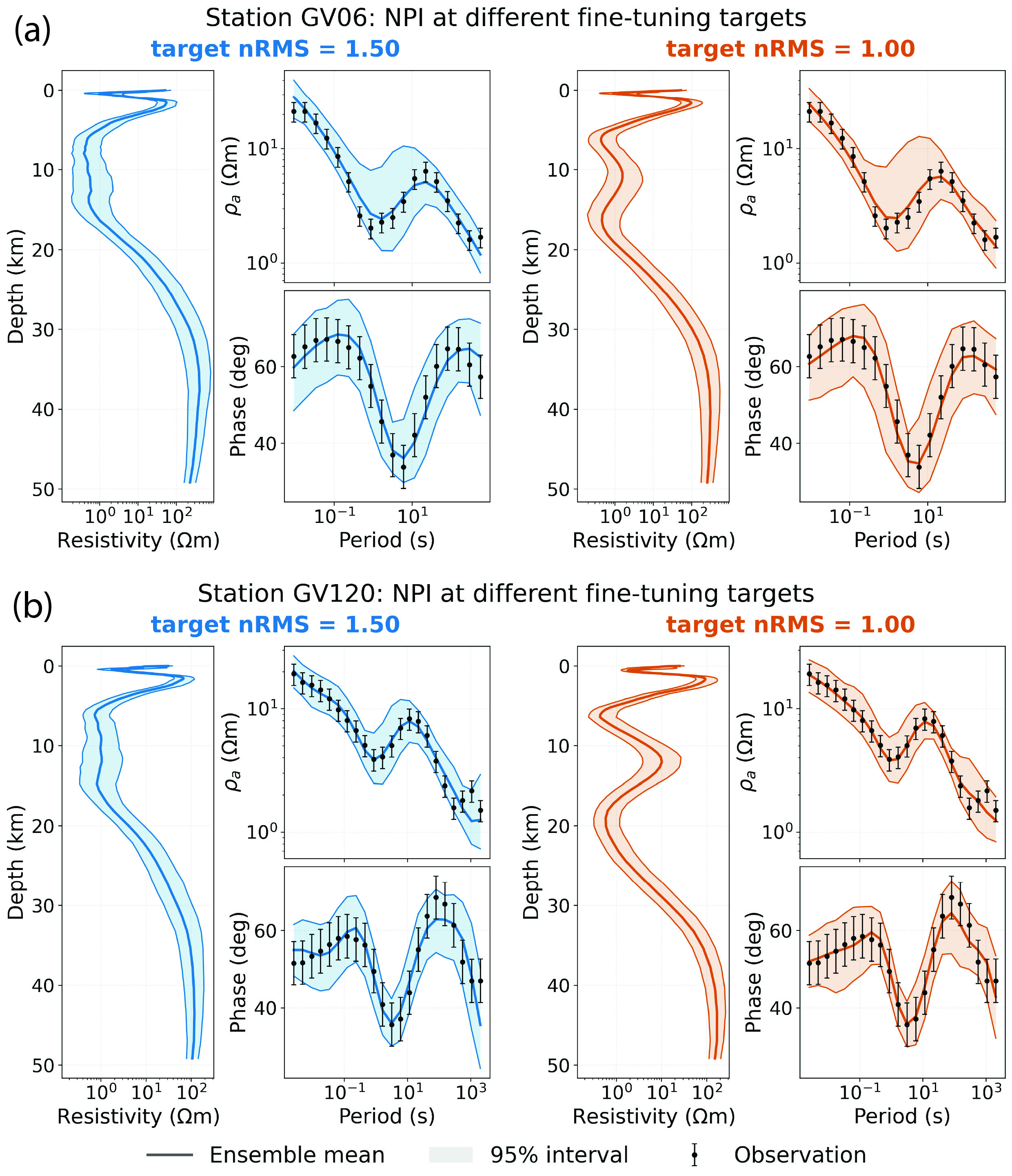}
\caption{
    Fine-tuning target sensitivity test. (a) Station GV06 and (b) station GV120, each at target nRMS = 1.50 (left, blue) and target nRMS = 1.00 (right, orange). In each, the left panel shows the resistivity ensemble mean and 95\% interval, and the right panels compare forward-modeled apparent resistivity and phase against the observed data. Relaxing the target to 1.50 suppresses internal resistivity structure within the conductive zone, acting similarly to stronger regularization, though through early stopping rather than a loss weight, while the large-scale conductive feature persists under either target.
    }
\label{fig:nrms_target_sensitivity}
\end{figure}

\section*{Appendix F: Phase Tensor Analysis and Dimensionality Diagnostics}
\label{app:phase_tensors}
Figure~\ref{fig:phase_tensors} shows phase tensor maps at six representative periods, providing dimensionality diagnostics that contextualize applying a 1D framework to field data that may contain higher-dimensional structure. Each ellipse represents the MT phase tensor \cite{caldwell_2004}, with color giving the skew angle $\beta$, where small $|\beta|$ is consistent with approximately 1D or 2D structure, while larger $|\beta|$ indicates three-dimensionality or lateral conductivity gradients \cite{caldwell_2004, booker_2014}. Because the phase tensor is insensitive to galvanic distortion, $\beta$ is a robust dimensionality diagnostic independent of static shift, though it is a continuous measure of complexity rather than a strict classification.

Across the array, the ellipses change in shape, orientation, and skew with period. At shorter periods most stations show compact, weakly skewed ellipses, consistent with simpler structure at shallow depth. At longer periods, increased ellipticity, rotation, and larger $\beta$ at some stations indicate stronger 3D effects in the deeper crust. An independent WALDIM analysis \cite{marti_2009}, which evaluates dimensionality from impedance-tensor rotational invariants, is consistent with these observations, showing similar depth- and period-dependent transitions between approximately 1D, 2D, and more strongly 3D behavior.

These diagnostics are qualitative context for interpreting the real-data NPI results, not used in the inversion or quantitatively linked to the spatial distribution of NPI ensemble spread; they indicate where the 1D assumption is more or less appropriate, motivating caution at stations and periods where multidimensionality is significant.

\begin{figure}[!htbp]
\centering
\includegraphics[width=0.9\columnwidth]{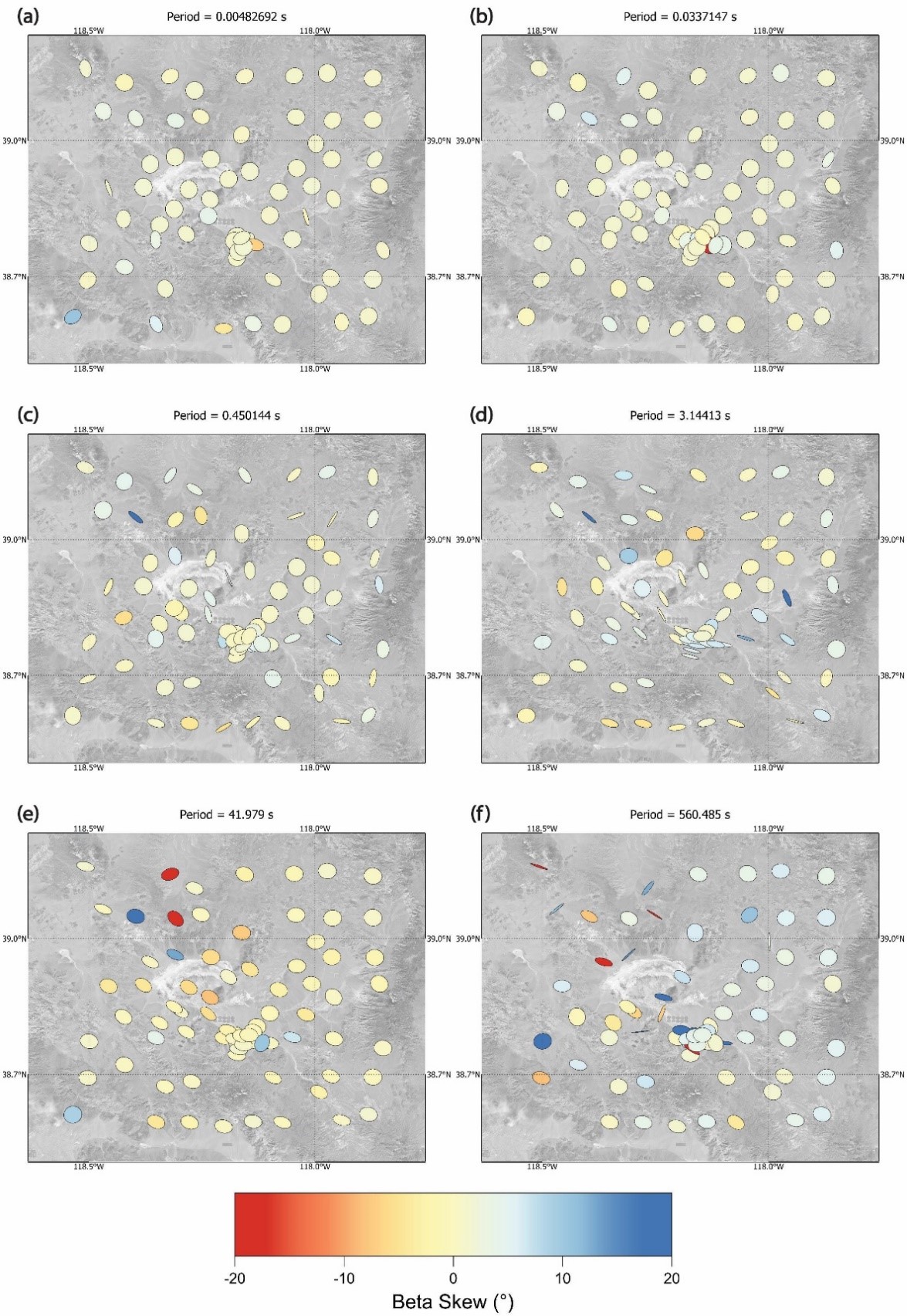}
\caption{
    Phase tensor maps for the Gabbs Valley MT dataset at six representative periods. Ellipse shape and orientation depict the phase tensor at each station \cite{caldwell_2004}, and color indicates the skew angle $\beta$, a measure of departure from 1D or 2D structure, with larger $|\beta|$ indicating stronger three-dimensional effects. Background is satellite imagery of the study area.
}
\label{fig:phase_tensors}
\end{figure}

\section*{Appendix G: Implementation Parameters and Computational Environment}
\label{app:implementation}
Table~\ref{tab:params} reports the parameters governing EnsCGP, the ResNet-1D pretraining shared by NPI and the end-to-end baseline, per-station fine-tuning, and RTO-TKO, to the level of detail needed for reproduction. Table~\ref{tab:timing} reports the computational environment and execution time underlying the comparison in Section~\ref{sec:NPI_real_data_application}, measured at station GV129.

EnsCGP forward models the conditioned ensemble in parallel across four worker processes. RTO-TKO draws form a serial Markov chain, since each draw's Gauss-Newton warm start is the previous draw's solution and the smoothing weight is updated by a sequential Gibbs step; within a single draw the Jacobian assembly parallelizes across frequency via a thread pool. We sample the regularization weight from its closed-form Gamma posterior (Table~\ref{tab:params}) rather than Blatter et al.'s (2022) \cite{blatter_part1_2022} uniform-prior, bounded-optimization TKO step, avoiding a second nested optimization per Gibbs iteration. Ensemble propagation through the NPI network batches members through a single forward pass, while the subsequent per-member forward-modeling step used to report the response ensemble runs as a serial loop.

\begin{table*}[!t]
\centering
\caption{Method Parameters}
\label{tab:params}
\begin{tabular}{p{2.3cm}p{4.3cm}p{10.5cm}}
\hline
\textbf{Component} & \textbf{Parameter} & \textbf{Value} \\
\hline
\multirow{3}{2.3cm}{EnsCGP}
 & Ensemble size $N$ & 500 \\
 & Localization & 500 nearest neighbors, cosine distance in data space \\
 & Observation noise model & $\sigma(\ln\rho_a)=0.10$, $\sigma(\phi)=2.86^\circ$ (5\% relative impedance error, Eq.~14) \\
\hline
\multirow{8}{2.3cm}{ResNet-1D pretraining}
 & Architecture & 2-channel input, conv stem (32), four residual blocks (32$\to$64$\to$128$\to$64$\to$32), global average pool, FC(32$\to$64$\to$89) \\
 & Optimizer & AdamW, learning rate $10^{-4}$, weight decay $10^{-4}$ \\
 & Loss & Huber, $\delta=1.0$ \\
 & Scheduler & ReduceLROnPlateau, patience 50, factor 0.5, minimum learning rate $10^{-6}$ \\
 & Batch size, max epochs, early-stop patience & 64, 1000, 200 \\
 & Cross-validation, seed & 5-fold, seed 0 \\
 & NPI input $\to$ target & EnsCGP ensemble-mean response $\to$ residual $\ln\rho_{true}-\ln\rho_{EnsCGP,mean}$ \\
 & End-to-end input $\to$ target & raw noisy response $\to$ full $\ln\rho_{true}$ profile \\
\hline
\multirow{6}{2.3cm}{Per-station fine-tuning (Eq.~8)}
 & Optimizer & Adam with gradient clipping (max norm 1.0), learning rate $10^{-4}$ \\
 & Stopping & maximum 3000 iterations, patience 20 epochs, minimum improvement $10^{-5}$ \\
 & $\alpha$ (Eq.~12) & 0.3, fixed \\
 & $\lambda_r$ (Eq.~13) & $10^{-2}$, fixed \\
 & $\lambda_s$ (Eq.~12) & adaptive: initial value $10^{2}$, multiplied by 1.2 or 0.8 each epoch depending on whether the current misfit sits above or below the target band ($\pm10^{-5}$), clipped to $[10, 10^{3}]$ \\
 & Target nRMS & 1.0 \\
\hline
\multirow{6}{2.3cm}{RTO-TKO}
 & Roughening blend $\alpha$ & 0.3 \\
 & $\lambda_s$ & initial 50.0 (10.0 in the standalone timing notebook), then Gibbs-sampled: $\lambda_s\mid m\sim\Gamma(a_0+\mathrm{rank}(L_s)/2,\,b_0+\tfrac12\lVert L_s m\rVert^2)$, $a_0=b_0=10^{-3}$ \\
 & $\lambda_r$ & $10^{-3}$, fixed \\
 & Data weights & identical to EnsCGP and NPI ($\sigma(\ln\rho_a)=0.10$, $\sigma(\phi)=2.86^\circ$) \\
 & Per-draw solve & damped Gauss--Newton, maximum 40 iterations, up to 30 backtracking steps, tolerance $10^{-7}$ \\
 & Draws & 500 kept, 50 burn-in \\
\hline
\end{tabular}
\end{table*}

\begin{table*}[!t]
\centering
\caption{Computational Environment and Inference Time at Station GV129}
\label{tab:timing}
\begin{tabular}{p{2.3cm}p{1.6cm}p{2.0cm}p{6.5cm}p{4.0cm}}
\hline
\textbf{Method} & \textbf{Time (s)} & \textbf{Ensemble} & \textbf{Per-member / per-draw cost} & \textbf{Execution} \\
\hline
EnsCGP & 1.03 & 500 & 0.00205 s per member & parallel, 4 worker processes \\
NPI & 11.96 & 500 & fine-tuning 11.01 s (independent of ensemble size); propagation and forward modeling 0.00190 s per member & fine-tuning single-threaded; propagation batched; per-member forward modeling serial \\
RTO-TKO & 640.07 & 500 (+50 burn-in) & 1.28 s per kept sample (1.16 s per draw including burn-in) & serial Markov chain; per-draw Jacobian assembly parallel, 16 threads \\
End-to-end$^\dagger$ & 5.19 & n/a & n/a, single point estimate & single-threaded \\
\hline
\multicolumn{5}{p{16.5cm}}{\footnotesize $^\dagger$ No ensemble; reported for completeness and not part of the EnsCGP, NPI, and RTO-TKO comparison in Section~\ref{sec:NPI_real_data_application}.}
\end{tabular}
\end{table*}

Hardware: AMD Ryzen 5 PRO 4650U (6 physical cores, 12 logical processors, 2.10~GHz base, 8.0~MB L3 cache), 7.9~GB RAM allocated to the WSL2 environment, no GPU. EnsCGP and RTO-TKO use NumPy and SciPy; NPI and the end-to-end baseline additionally use PyTorch (CPU only).

\section*{Appendix H: Capacity-Matched Ablation of the End-to-End Network}
\label{app:ablation}

As introduced in Section~\ref{sec:discussion_resnet}, we repeat the Table~\ref{tab:synthetic_summary} comparison with a reduced-capacity backbone applied identically to the end-to-end mapping and to the NPI residual correction (Equation~\ref{eq:residual_mapping}), so that architecture is held fixed across methods and any accuracy difference reflects the two-stage decomposition rather than capacity.

The reduced backbone (CompactNet1D) replaces the four residual blocks of Appendix~A with two plain convolutional layers (8 and 16 channels, kernel size 5), global average pooling, and a single linear layer mapping the 16-channel bottleneck directly to the 89-cell depth profile (2,257 parameters), a 75-fold reduction from the main architecture. The bottleneck width follows the effective dimensionality of the target: a singular value decomposition of the NPI residual field on out-of-fold validation data shows 9, 11, and 16 components explain 90\%, 95\%, and 99\% of depth-wise variance across the 89 cells, so a 16-channel bottleneck sits at approximately the 95\% point. Both arms otherwise follow the training protocol of Appendix~G (AdamW, Huber loss, \texttt{ReduceLROnPlateau}, 5-fold cross-validation).

For the end-to-end network we retain a dropout layer (rate 0.2) after the first convolutional layer, active during training and, for this ablation, also active at inference: keeping the network in evaluation mode but switching the dropout submodule back to training mode induces $N=300$ independent stochastic forward passes per example, forming a Monte Carlo dropout ensemble \cite{gal_2016_mc_dropout} as an approximate, architecture-preserving uncertainty estimate, a lighter-weight alternative to the diffusion or normalizing-flow redesign discussed in Section~\ref{sec:discussion_resnet}, applied post hoc to a network not trained to produce a distribution. The NPI arm uses the same reduced backbone without dropout at inference; its ensemble spread comes entirely from propagating the EnsCGP members through the deterministic network, rather than any stochastic sampling internal to the corrector.

We evaluate EnsCGP, NPI, and the end-to-end network with Monte Carlo dropout on 1,000 held-out synthetic examples drawn from the training prior (in-distribution) and a further 1,000 examples drawn under the homogeneous half-space prior of Appendix~C (out-of-distribution), which replaces the depth-structured mean and variance of the GMRF prior with a constant mean and a depth-independent standard deviation in $\log_{10}\rho$, using the accuracy, misfit, and coverage diagnostics of Section~\ref{sec:synt_test}. As in the main synthetic test, EnsCGP conditions each example once and the pretrained network is applied with parameters fixed; no per-station fine-tuning is used, matching the synthetic-testing protocol of Section~\ref{sec:testing} rather than the real-data protocol of Section~\ref{sec:real_adaptation}. The EnsCGP localization neighborhood, and correspondingly the ensemble size for EnsCGP and NPI, is $N=300$ in this ablation; all three methods compared here use matched ensemble or sample sizes, so the comparison remains internally consistent.

Figure~\ref{fig:ablation_examples} shows two representative examples, one in distribution and one under distribution shift, comparing EnsCGP, NPI, and the end-to-end network with Monte Carlo dropout. Table~\ref{tab:ablation} reports the results averaged over the full test sets.

\begin{figure}[!htbp]
\centering
\includegraphics[width=\columnwidth]{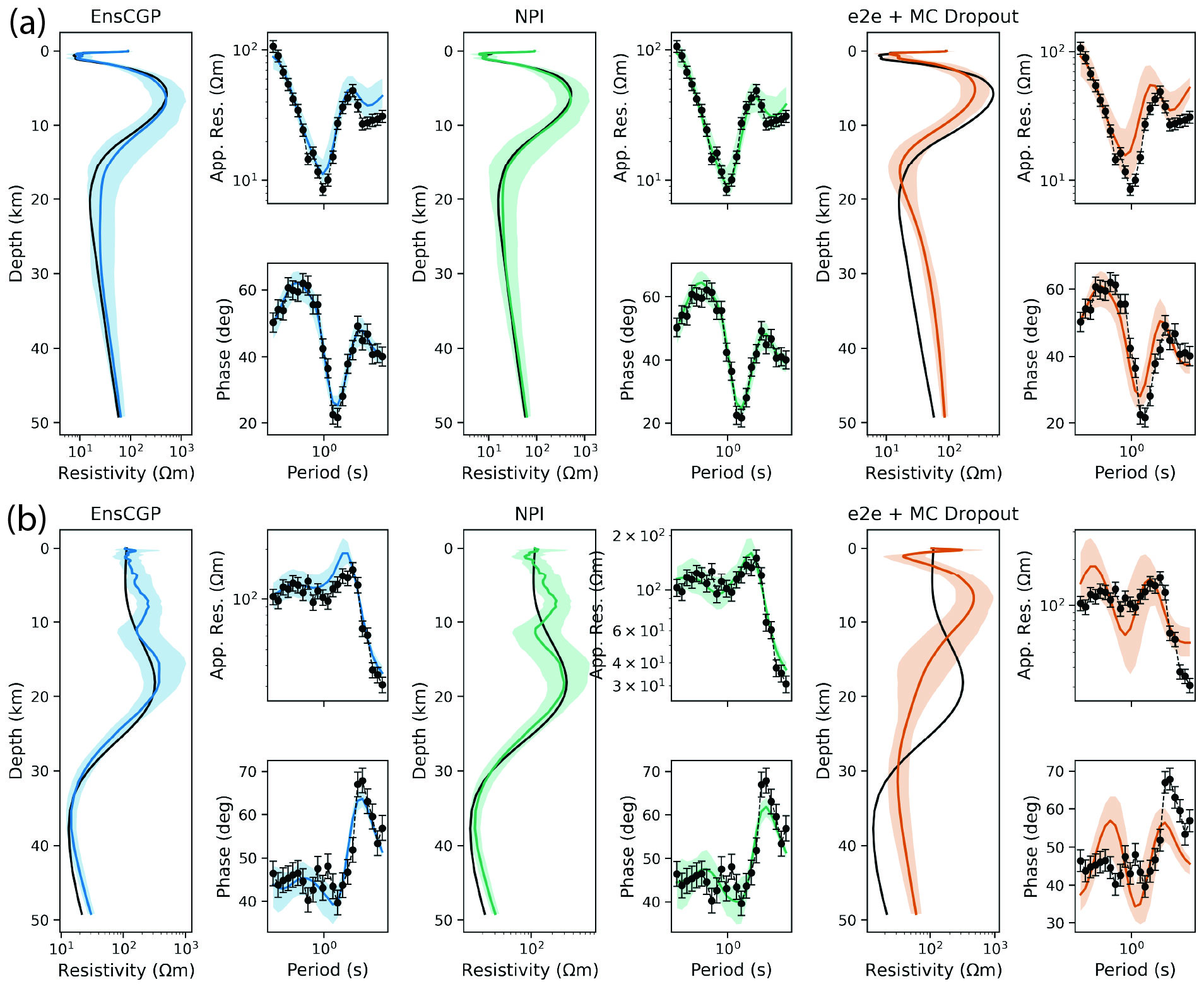}
\caption{
    Representative examples from the capacity-matched ablation, compared across EnsCGP, NPI, and the end-to-end network with Monte Carlo dropout (left to right within each row). For each method, the left panel shows the true resistivity model (black), ensemble mean (colored), and 95\% interval (shaded). Right panels show the corresponding forward-modeled apparent resistivity (top) and phase (bottom) against the observations (black, with error bars). (a) A representative in-distribution example. (b) A representative out-of-distribution example, drawn from the homogeneous half-space prior of Appendix~C.
}
\label{fig:ablation_examples}
\end{figure}

\begin{table*}[t]
\centering
\caption{Capacity-Matched Ablation: EnsCGP, NPI, and End-to-End with Monte Carlo Dropout}
\label{tab:ablation}
\begin{tabular}{llccccccc}
\toprule
& & \multicolumn{2}{c}{Accuracy} & \multicolumn{5}{c}{Ensemble} \\
\cmidrule(lr){3-4}\cmidrule(lr){5-9}
Setting & Method & RMS & nRMS & mean $U$ & 90\% & 95\% & Percentile & Spread/error \\
\midrule
\multirow{3}{*}{In distribution}
 & EnsCGP     & 0.124 & 1.200 & 1.79 & 88.3\% & 93.8\% & 42 & 0.98 \\
 & NPI        & 0.114 & 1.058 & 1.83 & 93.1\% & 96.8\% & 53 & 1.10 \\
 & End-to-end (MC dropout) & 0.191 & 1.798 & 1.58 & 59.2\% & 66.9\% & 47 & 0.52 \\
\midrule
\multirow{3}{*}{Out of distribution}
 & EnsCGP     & 0.196 & 1.608 & 1.68 & 72.7\% & 78.0\% & 44 & 0.55 \\
 & NPI        & 0.192 & 1.583 & 1.71 & 74.6\% & 79.9\% & 54 & 0.57 \\
 & End-to-end (MC dropout) & 0.346 & 2.874 & 1.83 & 46.2\% & 53.3\% & 45 & 0.36 \\
\bottomrule
\end{tabular}
\end{table*}

With capacity matched, the accuracy edge of the end-to-end network in Table~\ref{tab:synthetic_summary} disappears. NPI's RMS resistivity error and nRMS data misfit are each roughly 40\% lower than the end-to-end network's in distribution, widening under distribution shift, while EnsCGP alone already outperforms the end-to-end network on both metrics (Table~\ref{tab:ablation}).

NPI's coverage remains far closer to nominal than the end-to-end network's in both settings, with the truth well-centered in the ensemble throughout. The end-to-end network's Monte Carlo dropout ensemble reaches only 59--67\% coverage against 90\%/95\% nominal in distribution, roughly half the interval width reliable coverage requires, and degrades further under distribution shift, even though it too is centered near the true percentile. EnsCGP's own coverage also degrades under distribution shift, though less severely than the end-to-end network's.

}

\clearpage
\bibliographystyle{IEEEtran}
\bibliography{NPI}

\vfill

\end{document}